\documentclass{article}

\usepackage{iclr2027_conference,times}

\usepackage[utf8]{inputenc} 
\usepackage[T1]{fontenc}    
\usepackage{xurl}           
\usepackage{booktabs}       
\usepackage{array}          
\usepackage{amsfonts}       
\usepackage{amsmath}        
\usepackage{nicefrac}       
\usepackage{microtype}      
\usepackage{xcolor}         
\usepackage{graphicx}       
\usepackage{multirow}       
\usepackage{pifont}         
\usepackage{algorithm}      
\usepackage{algpseudocode}  
\usepackage{float}          
\usepackage{CJKutf8}        
\usepackage[most]{tcolorbox} 
\usepackage{placeins}     
\usepackage[hidelinks]{hyperref} 

\newcommand{\zh}[1]{\begin{CJK*}{UTF8}{gbsn}#1\end{CJK*}}
\newcommand{\zht}[1]{\begin{CJK*}{UTF8}{bsmi}#1\end{CJK*}}
\newtcolorbox{casebox}[1]{
  colback=gray!8,
  colframe=gray!35,
  boxrule=0.5pt,
  arc=3pt,
  left=5pt,
  right=5pt,
  top=4pt,
  bottom=4pt,
  title=#1,
  fonttitle=\bfseries,
  coltitle=black,
  before upper={\setlength{\parskip}{3pt}\setlength{\parindent}{0pt}}
}
\newtcolorbox{promptbox}[1]{
  breakable,
  colback=gray!8,
  colframe=gray!35,
  boxrule=0.5pt,
  arc=3pt,
  left=5pt,
  right=5pt,
  top=4pt,
  bottom=4pt,
  title=#1,
  fonttitle=\bfseries,
  coltitle=black,
  fontupper=\scriptsize,
  before skip=6pt,
  after skip=8pt,
  before upper={\raggedright\setlength{\parskip}{4pt}\setlength{\parindent}{0pt}}
}

\title{VisualNeedle: Benchmarking Active Visual Search in Information-Dense Scenes}

\author{
Jingru Chen\textsuperscript{1,2}\thanks{Equal contribution.}
\quad Yiming Liu\textsuperscript{1}\footnotemark[1]
\quad Mingtao Chen\textsuperscript{1}
\quad Sijie Chen\textsuperscript{1,3}
\\\bf
\quad Liang Yang\textsuperscript{1}
\quad Richeng Xuan\textsuperscript{1}
\quad Zhichao Hu\textsuperscript{1}
\quad Fanyang Lu\textsuperscript{1}\thanks{Correspondence to dylannlu@tencent.com.}
\\\normalfont\mdseries
\textsuperscript{1}Hunyuan, Tencent
\quad
\textsuperscript{2}Peking University
\quad
\textsuperscript{3}Zhejiang University}
\iclrfinalcopy

\begin{document}

\maketitle

\begin{abstract}
Frontier multimodal large language models (MLLMs) have been reported to
achieve over 90\% accuracy on fine-grained perception benchmarks. However,
such scores do not necessarily imply faithful use of visual evidence. Prior
studies have identified three shortcuts that inflate benchmark performance.
First, linguistic priors and lexical cues in questions often enable models to infer plausible
answers without seeing the image. Second, coarse global semantics from the
visual encoder can bypass fine-grained local details. Third, in some
``think-with-images'' benchmarks, corrupting the intermediate images returned
by visual tools barely affects the final answer.
These findings suggest that higher input resolution or larger question pools
alone do not elicit genuine active visual search. To address this, we introduce VisualNeedle, a challenging, information-dense, and fine-grained benchmark for scenes where critical evidence is spatially constrained to minute regions and not discernible at a glance. We further propose a counterfactual crop-black setting,
which replaces crops returned by tools with black images of the same size, to
test whether tool-enabled performance truly relies on intermediate visual
evidence.
We evaluate 9 prominent MLLMs across four settings: text-only, without tools, with tools, and crop-black. Text-only accuracy stays below 10\%, while accuracy without tools remains below 20\%. The best tool-enabled model reaches only 56.00\%, still trailing the 63.00\% human majority-vote accuracy. These results reveal persistent limitations in fine-grained visual search, while the crop-black ablation confirms that success on VisualNeedle hinges on genuine intermediate visual evidence.
\end{abstract}

\section{Introduction}
\label{sec:introduction}

Multimodal large language models (MLLMs) have made rapid progress on
visual understanding tasks, and the research community is increasingly
moving from passive image understanding, or ``thinking about images''~\citep{alayrac2022flamingo,li2023blip2,liu2023visual}, toward a more
active visual reasoning paradigm, or ``thinking with images''~\citep{su2025thinking,zheng2026deepeyes,zhang2025thyme,guo2025visualtoolbench,li2025tirbench,hong2025deepeyesv2}.
Under this paradigm, a model no longer receives the entire image once and
directly produces an answer; instead, it is expected to actively localize
relevant regions, acquire local visual evidence through operations such
as cropping, zooming, and code execution, and incorporate these
intermediate visual observations into subsequent reasoning.
Correspondingly, recent fine-grained perception and high-resolution
visual benchmarks have begun to design tasks around this active visual
capability: they ask models to identify small objects, read distant or
tiny text, judge local attributes, and understand subtle spatial
relations in complex images, thereby testing whether models can move from
a global scene to the local evidence needed for answering~\citep{yu2026finegrained,kanade2025doyouseeme,grover2026huemanity,feng2026reasonmap,li2025urbench}.
Leading systems now achieve over 90\% accuracy on several established
benchmarks, suggesting at first glance that fine-grained visual
perception, and even active visual search, may be approaching saturation.
This raises a more fundamental evaluation question: do these high scores
indicate that models can actively search for, localize, and use fine-grained
visual evidence?

This concern is not merely speculative. Three recent findings suggest that
benchmark accuracy can be decoupled from genuine visual evidence use.
First, language priors and lexical cues in the question can lead models to
produce plausible answers without seeing the image~\citep{asadi2026mirage}. Second, visual encoders may retain coarse global
semantics while missing fine local details, allowing models to rely on
scene-level gist instead of resolving the decisive clue~\citep{tong2024eyes}. Third, in ``think-with-images'' pipelines,
corrupting intermediate visual outputs can leave final decisions largely
unchanged, indicating that tool use does not necessarily imply reliance on
the returned visual evidence~\citep{liu2025faithfulness}. Together, these
failure modes make high accuracy an ambiguous signal: a benchmark may
appear saturated while still failing to test whether models actively
search for, localize, and use fine-grained visual evidence.

This makes active visual search a measurement problem, not merely a task
difficulty problem. Simply increasing image resolution or adding more
questions does not guarantee that models must engage with local evidence.
A benchmark for this setting should instead be constructed so that
shortcut solutions fail by design: the question alone should be
insufficient, a one-shot global view should not reveal the answer, and
tool-enabled success should depend on the visual content returned during
interaction.

\begin{figure*}[t]
  \centering
  \includegraphics[width=\textwidth]{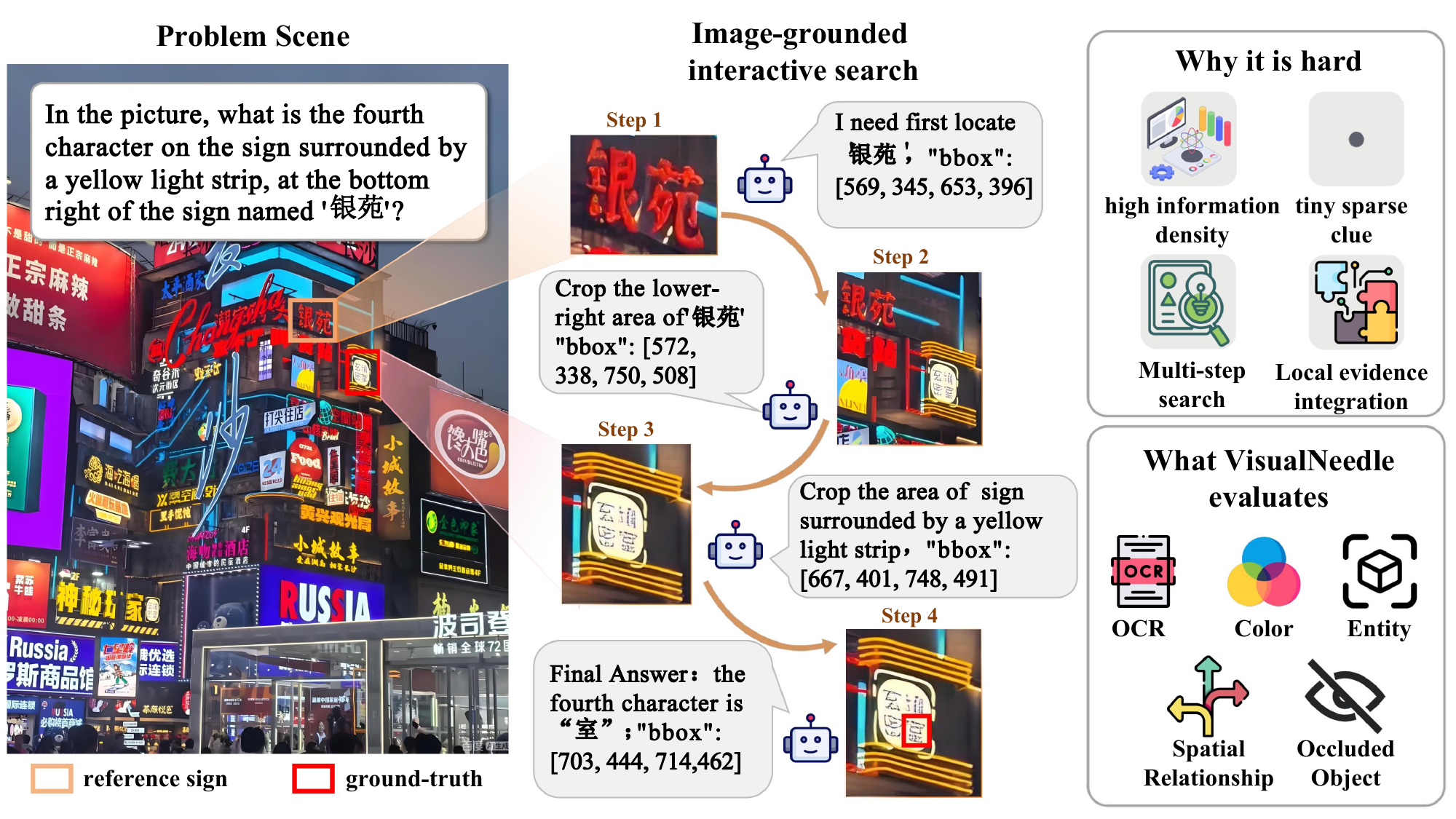}
  \caption{
  Overview of VisualNeedle and the interactive search workflow.
  Left: an example query in an information-dense scene.
  Middle: the step-by-step search process through sequential crops.
  Right: the core challenges and evaluated abilities.
  }
  \label{fig:overview}
\end{figure*}

Guided by this principle, we introduce \textbf{VisualNeedle}
(Figure~\ref{fig:overview}), a 300-question benchmark for active visual
search in information-dense scenes, including urban street views, dense
documents, shelves, bookcases, and maps. Each question satisfies an
answer-not-visible-at-a-glance criterion and is anchored to a small but
decisive local visual clue. The
benchmark spans five complementary abilities: OCR recognition, color recognition,
entity recognition, spatial relationship, and occluded object
recognition, all requiring models to move from a cluttered global scene to
specific local evidence before answering. This design makes each instance
depend on finding the right local evidence rather than recognizing the
overall scene type or exploiting answer regularities.

Beyond static evaluation, VisualNeedle directly tests whether models
use the evidence discovered during search. We formulate the task as a
multi-step tool-use process and introduce the crop-black setting, which
replaces returned crop images with all-black images of the same size
while preserving the interaction protocol and tool schema. Comparing
this setting with standard Crop-based evaluation reveals whether
tool-enabled gains depend on the returned visual content. Because the
ablation preserves the tool interface and returned-image dimensions,
the resulting gap reveals reliance on task-relevant visual features during reasoning.

We evaluate 9 mainstream MLLMs under four settings: text-only tests
question-level priors; the without-tools setting tests one-shot full-image
perception; the with-tools setting provides real tool-based local evidence; and the
crop-black setting preserves the tool-use protocol while replacing Crop
outputs with all-black patches. The results validate the benchmark design.
Text-only accuracy stays below 10\%, showing that
question-level priors are insufficient; accuracy without tools stays below 20\%,
showing that one-shot global perception is also insufficient. With tools,
the strongest model, Gemini 3.1 Pro, reaches 56.00\%, still below the human
majority accuracy of 63.00\%. The crop-black setting further shows
that successful performance on VisualNeedle depends on real intermediate
visual evidence. Together, these results test the three intended failure modes:
language priors, one-shot global perception, and tool-output invariance.
Additional trajectory and tool-use analyses reveal that models differ not
only in whether they call tools, but also in whether they can localize
evidence and integrate it reliably.

These findings demonstrate that the active visual search capability of
current MLLMs in information-dense scenes remains fundamentally limited. Our contributions are threefold:
\begin{enumerate}
  \item \textbf{We identify shortcut-driven mismeasurement in current
  fine-grained visual benchmarks and translate it into benchmark design
  constraints.} VisualNeedle is constructed so that language priors,
  coarse global semantics, and tool-output invariance are insufficient for
  reliable success.

  \item \textbf{We build VisualNeedle, a 300-question benchmark for active
  visual search in information-dense scenes.} The benchmark covers
  five fine-grained abilities and requires models to locate small local
  evidence before answering.

  \item \textbf{We introduce the crop-black setting to test
  intermediate visual evidence dependence.} The results show that
  success on VisualNeedle depends on the actual returned crops, exposing
  limitations in current MLLMs' active visual search behavior.
\end{enumerate}

\section{Related Work}

Our work builds upon and extends research in visual-language benchmarks,
high-resolution perception, and visual tool use. We organize related work along
three dimensions that highlight the unique contributions of VisualNeedle.

\subsection{Visual-Language Benchmarks and Their Limitations}

Early vision-language benchmarks such as MME~\citep{fu2023mme},
MMBench~\citep{liu2023mmbench}, and MMMU~\citep{yue2023mmmu} focused
primarily on general visual question answering, where models typically
receive the whole image once and produce a final answer. Recent benchmarks
including MME-RealWorld~\citep{zhang2024mmerealworld} and
RealX-Bench~\citep{hong2025deepeyesv2} move toward harder real-world
visual understanding by increasing image realism, resolution, and scene
complexity. These benchmarks have been important for tracking the broad
progress of MLLMs, but their evaluation remains largely final-answer
based. As a result, high accuracy alone may not reveal whether a model
grounds its answer in the necessary visual evidence.

\subsection{High-Resolution and Fine-Grained Perception Benchmarks}

Recent benchmarks have examined high-resolution and fine-grained
perception from complementary perspectives. HR-Bench~\citep{wang2024dc2}
studies information loss and perceptual hallucination caused by image
resizing, while V* Bench~\citep{wu2024vstar} introduces active visual
search by requiring models to locate target objects, and
VisualOverload~\citep{gavrikov2025visualoverload}
further stresses MLLMs with extremely dense visual scenes. VisualProbe-Hard~\citep{lai2026mini} evaluates visual search for small targets amid numerous distractors in high-resolution images. More recent
benchmarks broaden this direction to fine-grained reasoning over transit
maps~\citep{feng2026reasonmap}, multidimensional visual perception~\citep{kanade2025doyouseeme},
fine-grained pattern recognition~\citep{grover2026huemanity}, general
fine-grained image tasks~\citep{yu2026finegrained}, and multi-hop
reasoning over ultra-high-resolution images~\citep{li2025urbench}.

Together, these works show that resolution, visual density, local detail,
and spatial grounding are central to evaluating fine-grained perception.
However, they do not simultaneously require shortcut-resistant question
design, diverse information-dense real-world scenes, and a
diagnostic test of whether tool-enabled success depends on returned
intermediate visual content. VisualNeedle is designed to combine these
requirements in a single benchmark.

\subsection{Visual Tool Use and Interactive Evaluation}

The growing capability of visual agents has spurred research on tool use
and interactive evaluation. DeepEyes~\citep{zheng2026deepeyes},
Thyme~\citep{zhang2025thyme}, and ACTIVE-o3~\citep{zhu2026activeo} use
reinforcement learning to train models to call tools such as Crop or to
select zoom-in regions, improving performance on fine-grained tasks.
VisualToolBench~\citep{guo2025visualtoolbench} constructs open-ended tasks
for general visual tool-use evaluation. TIR-Bench~\citep{li2025tirbench}
extends evaluation to multiple task types involving tool use and
image-processing reasoning. VTC-Bench~\citep{zhu2026vtcbench} provides
OpenCV tools for evaluating composition and planning in multi-step visual
tool chains.

These works establish tool use as an important capability for MLLMs, but
successful tool calling does not by itself demonstrate reliance on the
visual content returned by tools. VisualNeedle therefore focuses on a
narrower but more diagnostic question: in dense local visual search, do
models use the intermediate visual evidence obtained through interaction?
The crop-black setting directly targets this question by preserving the
tool interface and interaction protocol while removing the returned visual
content.

\section{VisualNeedle Benchmark}

VisualNeedle is designed as a constructive response to the three shortcuts
identified above: language priors, coarse global semantics, and weak
dependence on intermediate visual evidence. Rather than treating these
shortcuts as post hoc diagnostic observations, we translate them into
three benchmark requirements: the question alone should be insufficient,
a one-shot global view should be insufficient, and tool-enabled success
should depend on the visual content returned during interaction. The task
definition, curation pipeline, and the crop-black setting are built
around these requirements. The following subsections describe the resulting
task protocol, data construction process, and dataset characteristics.

\subsection{Task Definition}

VisualNeedle formulates fine-grained visual perception as a sequential
visual search problem rather than a one-shot VQA problem. Given a
high-resolution image $I$ and a question $Q$, the model starts from the
full-image observation $o_0=I$ and repeatedly chooses an action
$a_t \in \{\textbf{Crop}(bbox_t), \textbf{Answer}(y_t)\}$: \textbf{Crop}
returns a localized visual observation, while \textbf{Answer} terminates
the interaction and produces the final answer. This protocol also enables
the crop-black setting, where cropped observations are replaced
with all-black patches of the same size while the interaction protocol is
preserved. This tests whether the model's answer depends on intermediate
visual evidence rather than merely on the act of calling tools; the
complete stepwise protocol is specified in Appendix
Algorithm~\ref{alg:task_definition}.

\subsection{Design Constraints and Curation Pipeline}
\label{sec:curation_pipeline}

VisualNeedle is constructed through a constraint-driven curation protocol
rather than standard image-question sampling. Each sample is required to
satisfy three anti-shortcut constraints that directly instantiate the
requirements above. First, \textit{question-alone insufficiency}: the
answer should not be recoverable from the wording of the question or its
lexical cues. Second, \textit{one-shot global-view insufficiency}: the
decisive evidence should not be reliably captured from the downsampled
whole-image view. Third, \textit{returned-evidence dependence}: the local
evidence discovered during interaction should be visually inspectable, so
that tool-enabled performance can be tested against the crop-black setting.
These constraints guide the following
four-stage pipeline.

\begin{enumerate}
\item \textbf{Source image selection.}
We first collect information-dense images from diverse real-world
scenes, including complex street views, dense document scans, shelves
packed with products, bookcases, and panoramic maps. Image selection is
not driven by resolution alone. Instead, we prioritize scenes with rich
local structure, many candidate entities, diverse object categories, and
dense segmentation regions, so that coarse global semantics are
insufficient for reliable answering.

\item \textbf{Needle mining.}
During data construction, dataset curators inspect full-resolution images
to identify small but recognizable visual evidence, which we refer to as
``needles.'' A needle may be distant text on a storefront sign, the title of
a book on a shelf, a local attribute of a person, or the category of a
partially occluded object. This step anchors each question to concrete local
evidence and turns the task from passive image understanding into active
visual search.

\item \textbf{Question construction.}
Around each mined needle, the data-construction team writes questions
covering five core ability categories: OCR recognition, color recognition,
entity recognition, spatial relationship, and occluded object
recognition. The question is written so that lexical cues alone are
insufficient to infer the answer, while the correct answer becomes clear once
the target region is localized and inspected. Many questions require
progressive search and multiple perceptual operations, such as first
locating a person in a crowded scene and then judging the color of that
person's clothing.

\item \textbf{Annotation and verification.}
Each candidate sample undergoes multi-round cross-checking by different
members of the data-construction and verification team. The verification
process checks that the wording is unambiguous, the answer is unique, the
target evidence is visible under appropriate local observation, and the
answer is not visible at a glance from the original image scale. This
construction-stage verification is separate from the human-baseline
annotation used in our evaluation: the 12 external human-baseline annotators
did not participate in image collection, question authoring, bounding-box
drawing, or quality filtering, and they answered the final evaluation items
independently without access to ground-truth answers or construction-time
metadata.
\end{enumerate}

\subsection{Dataset Composition and Diagnostic Coverage}

\begin{figure*}[t]
  \centering
  \includegraphics[width=0.98\textwidth]{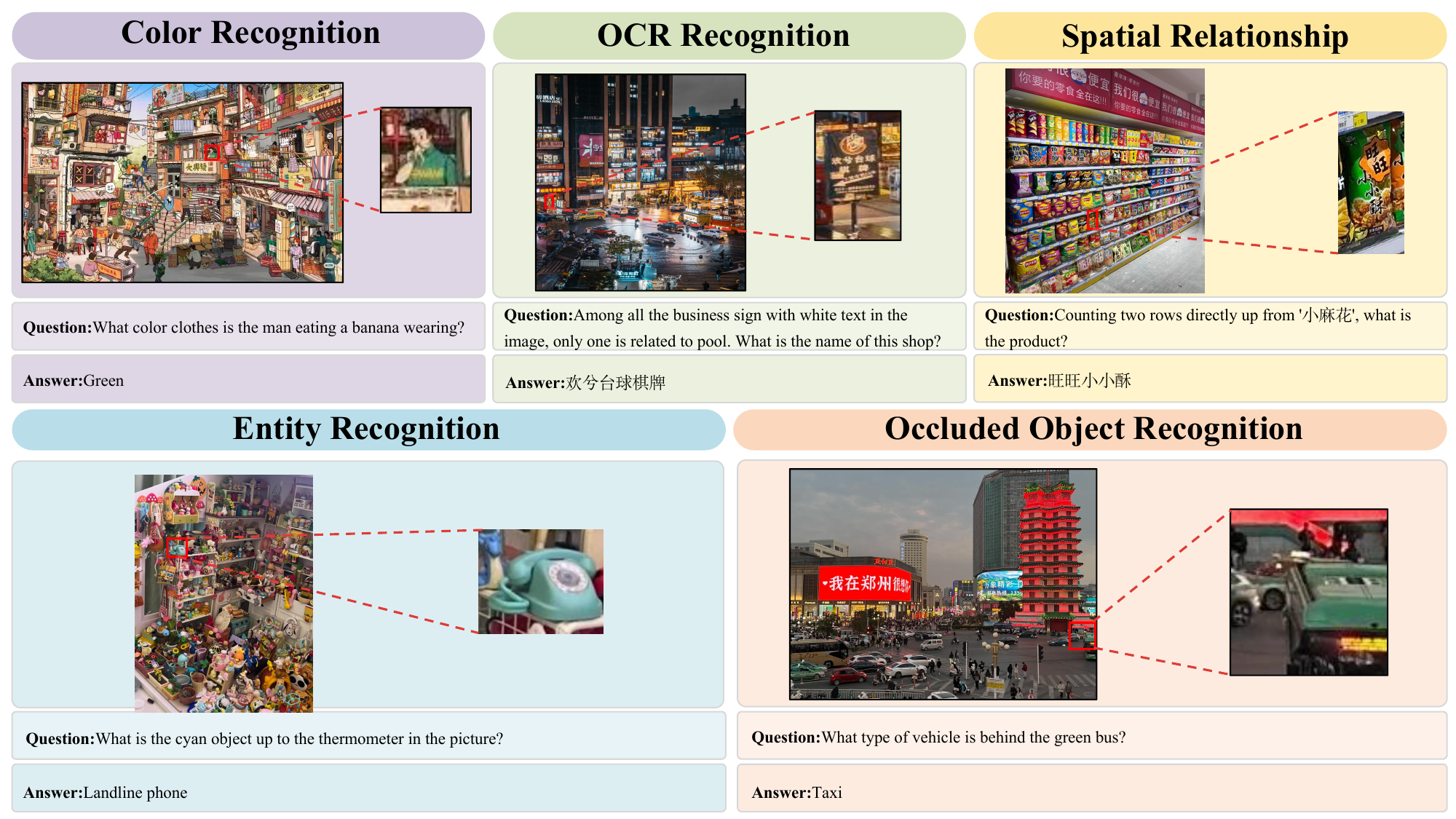}
  \caption{
  Examples from VisualNeedle showing how questions require moving from an
  information-dense global scene to a small piece of local visual evidence. The examples probe complementary failure 
 modes, including color recognition, OCR recognition, spatial relationship, entity recognition, and
occluded object recognition. 
  }
  \label{examples}
\end{figure*}

VisualNeedle contains 300 curated questions from information-dense
scenes, including street views, documents, shelves, bookcases, and maps.
Its examples span five fine-grained ability categories: OCR recognition,
color recognition, entity recognition, spatial relationship, and
occluded object recognition. Representative examples are shown in Figure~\ref{examples}. Rather than serving as broad topical
coverage alone, these categories are selected to expose distinct failure
modes in active visual search.

OCR recognition probes tiny text localization and reading;
color recognition tests localized attribute grounding; entity recognition
measures target localization in cluttered scenes; spatial relationship requires local relational reasoning after search; and occluded
object recognition evaluates inference from incomplete local evidence.
Together, these categories make each example a diagnostic probe of whether
the model can move from a global scene to the small visual clue needed for
the answer.

The target-region area distribution is reported in
Appendix~\ref{app:target_region_size}. Across the 300 questions, 99.00\%
of annotated target boxes occupy no more than 1.00\% of the original
full-image area.

Table~\ref{tab:benchmark_complexity} further clarifies where the
difficulty of VisualNeedle comes from. The benchmark is not designed
to be difficult merely because it has more questions or the highest
raw resolution. Instead, its difficulty comes from dense local visual
structure: many candidate entities, diverse object categories, fine
segmentation regions, and high visual complexity within each image.
Although VisualNeedle is not the highest-resolution benchmark, it has
substantially denser local structure, as reflected by
DINO-X~\citep{ren2024dinox} entity and category
counts, SAM~\citep{kirillov2023sam} segmentation regions, and the
Complexity in Complexity metric~\citep{saritas2025complexity}.
This supports the central goal of evaluating active visual search in
scenes where coarse global perception is insufficient.

\begin{table*}[t]
  \centering
  \footnotesize
  \setlength{\tabcolsep}{5pt}
  \caption{
  Comparison between VisualNeedle and existing related benchmarks in
  resolution and complexity metrics.
  }
  \label{tab:benchmark_complexity}
  \begin{tabular}{lcccccc}
    \toprule
    Benchmark & \shortstack{\#Evaluation\\Instances}
    & \shortstack{Resolution\\(MP)}
    & \shortstack{DINO-X\\Entities}
    & \shortstack{DINO-X\\Categories}
    & \shortstack{SAM\\Segments}
    & \shortstack{Complexity in\\Complexity} \\
    \midrule
    V* Bench & 191 & 4.05 & 79.86 & 17.76 & 169.46 & 0.49 \\
    HR-Bench (8K) & 800 & 39.44 & 53.11 & 10.40 & 140.62 & 0.40 \\
    TIR-Bench & 1215 & 3.61 & 35.26 & 8.76 & 186.73 & 0.37 \\
    VisualToolBench & 1204 & 3.55 & 42.94 & 6.38 & 159.16 & 0.39 \\
    VisualOverload & 2720 & 7.78 & 129.48 & 23.29 & 187.02 & 0.52 \\
    MME-RealWorld & 29,429 & 11.47 & 86.67 & 13.92 & 148.80 & 0.42 \\
    VisualProbe-Hard & 106 & 20.88 & 110.03 & 18.41 & 224.58 & 0.53 \\
    VTC-Bench & 680 & 1.36 & 26.31 & 4.71 & 102.08 & 0.27 \\
    \textbf{VisualNeedle (ours)}
    & \textbf{300} & \textbf{7.46} & \textbf{122.79}
    & \textbf{24.55} & \textbf{352.17} & \textbf{0.73} \\
    \bottomrule
  \end{tabular}
\end{table*}

\section{Experiments}

\subsection{Experimental Setup}

We evaluate 9 mainstream models on all 300 VisualNeedle questions,
covering closed-source commercial models such as Gemini 3.1 Pro and
GPT-5.4 as well as open-source models such as Qwen3-VL-235B-A22B-Thinking. Following
the benchmark design constraints, we consider four evaluation protocols:
\textit{text-only}, where the model receives only the question;
\textit{without tools} (w/o tools), where the model receives a one-shot global view of the
image and must answer directly; \textit{with tools} (w/ tools), where the model can
call the \textbf{Crop} tool to obtain high-resolution local observations;
and the crop-black setting, where the interaction protocol is preserved but
returned crops are replaced with all-black images of the same size. All models
are evaluated with the same evaluation harness to ensure consistency in
multi-step interaction and answer judgment. We repeat the main experiments without and
with tools three times under the same setting and report the
mean values in Table~\ref{tab:main_results}; the corresponding run-to-run
variance and 95\% confidence intervals are reported in
Appendix~\ref{app:repeat_variance}. The text-only and crop-black
diagnostics are run once per model and reported in
Table~\ref{tab:cropblackablation}.

\subsection{Main Results}

\begin{table*}[t]
  \centering
  \scriptsize
  \setlength{\tabcolsep}{3.5pt}
  \caption{
  Main experimental results on VisualNeedle, averaged over three repeated runs under the same setting.
  $\Delta$ denotes the absolute accuracy improvement in percentage points after tools are introduced.
  Avg. Tool Calls denotes the mean number of Crop calls per question.
  Mean Total Tool Calls denotes the total number of Crop calls over the 300-question evaluation split, averaged across the three runs.
  The lower part of the table reports human-baseline results.
  }
  \label{tab:main_results}
  \resizebox{\textwidth}{!}{
  \begin{tabular}{lccccc}
    \toprule
    Model
    & \shortstack{w/ tools\\Acc. (\%)}
    & \shortstack{w/o tools\\Acc. (\%)}
    & \shortstack{$\Delta$\\(pp)}
    & \shortstack{Avg. Tool\\Calls}
    & \shortstack{Mean Total\\Tool Calls} \\
    \midrule
    Gemini 3.1 Pro~\citep{google2026gemini31pro}
    & 56.00 & 16.33 & 39.67 & 3.72 & 1117.33 \\

    Gemini 3 Flash~\citep{google2025gemini3flash}
    & 48.00 & 15.11 & 32.89 & 3.91 & 1174.00 \\

    Doubao-Seed-2.0-Pro~\citep{seed2026seed2}
    & 36.78 & 13.11 & 23.67 & 6.37 & 1912.33 \\

    Doubao-Seed-1.8~\citep{seed2026seed1}
    & 31.78 & 17.89 & 13.89 & 4.78 & 1434.67 \\

    GPT-5.4~\citep{openai2026gpt54}
    & 31.22 & 14.33 & 16.89 & 11.07 & 3322.00 \\

    GPT-5.2~\citep{openai2025gpt52}
    & 28.44 & 10.56 & 17.89 & 9.19 & 2756.00 \\

    Qwen3.5-Plus~\citep{qwen2026qwen35plus}
    & 22.11 & 15.00 & 7.11 & 3.49 & 1046.00 \\

    Kimi K2.5~\citep{moonshot2026kimi25}
    & 19.33 & 10.78 & 8.56 & 6.58 & 1974.33 \\

    Qwen3-VL-235B-A22B-Thinking~\citep{qwen2026qwen3vl235b}
    & 11.44 & 8.33 & 3.11 & 1.58 & 473.33 \\

    \midrule
    \multicolumn{6}{l}{\textbf{Human Baseline}} \\
    \midrule
    Metric
    & \shortstack{Correct\\Count}
    & Accuracy (\%)
    & \multicolumn{3}{c}{--} \\

    Majority Vote@3
    & 189/300
    & 63.00
    & \multicolumn{3}{c}{--} \\

    Pass@3
    & 224/300
    & 74.67
    & \multicolumn{3}{c}{--} \\
    \bottomrule
  \end{tabular}
  }
\end{table*}

\textbf{Active visual search remains challenging for current MLLMs.}
The main results support three observations. First, one-shot global
perception fails on VisualNeedle. Without tools, all models remain
below 20\% accuracy, with the best model reaching only 17.89\%. This
confirms that the decisive evidence in VisualNeedle is usually not
recoverable from a downsampled whole-image view, and that answering from
global scene semantics is insufficient.

Second, visual tools help substantially but do not close the human gap.
With Crop, stronger models obtain large gains: Gemini 3.1 Pro improves by
39.67 percentage points and reaches 56.00\%, while Gemini 3 Flash reaches
48.00\%. However, even the strongest model remains below the human majority
accuracy of 63.00\% and far below the human Pass@3 accuracy of 74.67\%.
This suggests that tools can expose missing local details, but current
models still struggle to search for and integrate fine-grained evidence as
reliably as humans.

Third, tool calling is not equivalent to effective evidence use. Some
models call tools frequently but obtain limited gains: GPT-5.4 and GPT-5.2
average 11.07 and 9.19 Crop calls, respectively, yet remain far below the
leading Gemini models. Conversely, Qwen3-VL-235B-A22B-Thinking calls tools
infrequently and improves by only 3.11 points, suggesting premature
stopping or ineffective localization. These results indicate that the core
challenge is not merely whether a model can invoke Crop, but whether it can
localize the relevant region and use the returned evidence reliably.

\subsection{Intermediate Visual Evidence Dependence}

This diagnostic directly tests the third design constraint:
\textit{returned-evidence dependence}. A tool-enabled model should improve
because it uses the visual content returned by Crop, not simply because it
has access to the tool interface and goes through the interaction. We therefore contrast standard evaluation with tools
against two diagnostic settings. In the \textit{text-only} setting,
the model receives only the question, which probes whether language priors
and lexical cues are sufficient. In the crop-black setting, the
model can still follow the same tool-use protocol, but every Crop call
returns an all-black patch instead of the real cropped region. This removes
intermediate visual evidence while preserving the interaction format.

\begin{table*}[t]
  \centering
  \footnotesize
  \setlength{\tabcolsep}{4.5pt}
  \caption{
  Accuracy under the text-only, crop-black, and w/ tools settings.
  Text-only denotes accuracy when only the question is provided;
  crop-black reports accuracy when Crop returns all-black patches;
  w/ tools denotes accuracy when Crop returns real cropped images.
  The VisualNeedle w/ tools column is the three-run mean from
  Table~\ref{tab:main_results}; the text-only and crop-black columns and all
  HR-Bench and V* Bench columns are from a single run.
  }
  \label{tab:cropblackablation}
  \resizebox{\textwidth}{!}{
  \begin{tabular}{lccccccccc}
    \toprule
    \multirow{2}{*}{Model} &
    \multicolumn{3}{c}{VisualNeedle (300)} &
    \multicolumn{3}{c}{HR-Bench} &
    \multicolumn{3}{c}{V* Bench} \\
    \cmidrule(lr){2-4} \cmidrule(lr){5-7} \cmidrule(lr){8-10}
    & text-only & crop-black & w/ tools
    & text-only & crop-black & w/ tools
    & text-only & crop-black & w/ tools \\
    \midrule
    Gemini 3.1 Pro              & 7.67 & 12.00 & 56.00 & 47.00 & 71.00 & 88.00 & 46.07 & 84.29 & 97.38 \\
    Gemini 3 Flash              & 6.67 & 13.33 & 48.00 & 48.50 & 84.00 & 86.00 & 40.31 & 87.43 & 96.86 \\
    Doubao-Seed-2.0-Pro         & 8.00 & 13.33 & 36.78 & 52.00 & 82.00 & 93.00 & 31.94 & 86.91 & 93.72 \\
    Doubao-Seed-1.8             & 8.33 & 15.33 & 31.78 & 47.00 & 86.50 & 92.00 & 41.36 & 86.39 & 91.62 \\
    GPT-5.4                     & 6.33 &  8.33 & 31.22 & 43.50 & 64.50 & 94.50 & 42.41 & 75.92 & 97.38 \\
    GPT-5.2                     & 1.00 &  7.00 & 28.44 & 16.00 & 70.00 & 87.50 & 38.74 & 76.44 & 95.81 \\
    Qwen3.5-Plus                & 4.00 &  6.67 & 22.11 & 47.50 & 38.50 & 90.50 & 33.51 & 80.63 & 93.72 \\
    Kimi K2.5                   & 6.67 & 10.00 & 19.33 & 42.50 & 36.50 & 86.00 & 36.13 & 82.72 & 93.19 \\
    Qwen3-VL-235B-A22B-Thinking & 5.67 &  7.67 & 11.44 & 39.50 & 35.00 & 84.00 & 34.03 & 85.34 & 92.67 \\
    \bottomrule
  \end{tabular}}
\end{table*}

Table~\ref{tab:cropblackablation} shows that VisualNeedle strongly
depends on intermediate visual content. For all models, accuracy in the
crop-black setting remains close to text-only accuracy, while
real crops bring large gains for stronger models; for example, Gemini 3.1
Pro rises from 12.00\% under the crop-black setting to 56.00\% with real
crops, and Gemini 3 Flash rises from 13.33\% to 48.00\%. These gaps show
that tool gains on VisualNeedle come from using the returned visual
evidence, not merely from having access to the tool interface.

The contrast with HR-Bench and V* Bench is substantial. On V* Bench, all models stay
within 75.92\%--87.43\% even when crop content is replaced by black
patches. HR-Bench is more mixed, but several strong models also retain high
accuracy in the crop-black setting, such as Gemini 3 Flash at 84.00\%,
Doubao-Seed-2.0-Pro at 82.00\%, and Doubao-Seed-1.8 at 86.50\%. These
results suggest that prior benchmarks can often still be solved from
global context, question cues, or the act of interacting with tools, whereas
VisualNeedle makes the intermediate visual evidence necessary for
tool-enabled success.

\subsection{Tool Efficiency and Stability}

\begin{figure*}[t]
  \centering
  \begin{minipage}[t]{0.48\textwidth}
    \centering
    \includegraphics[width=\textwidth]{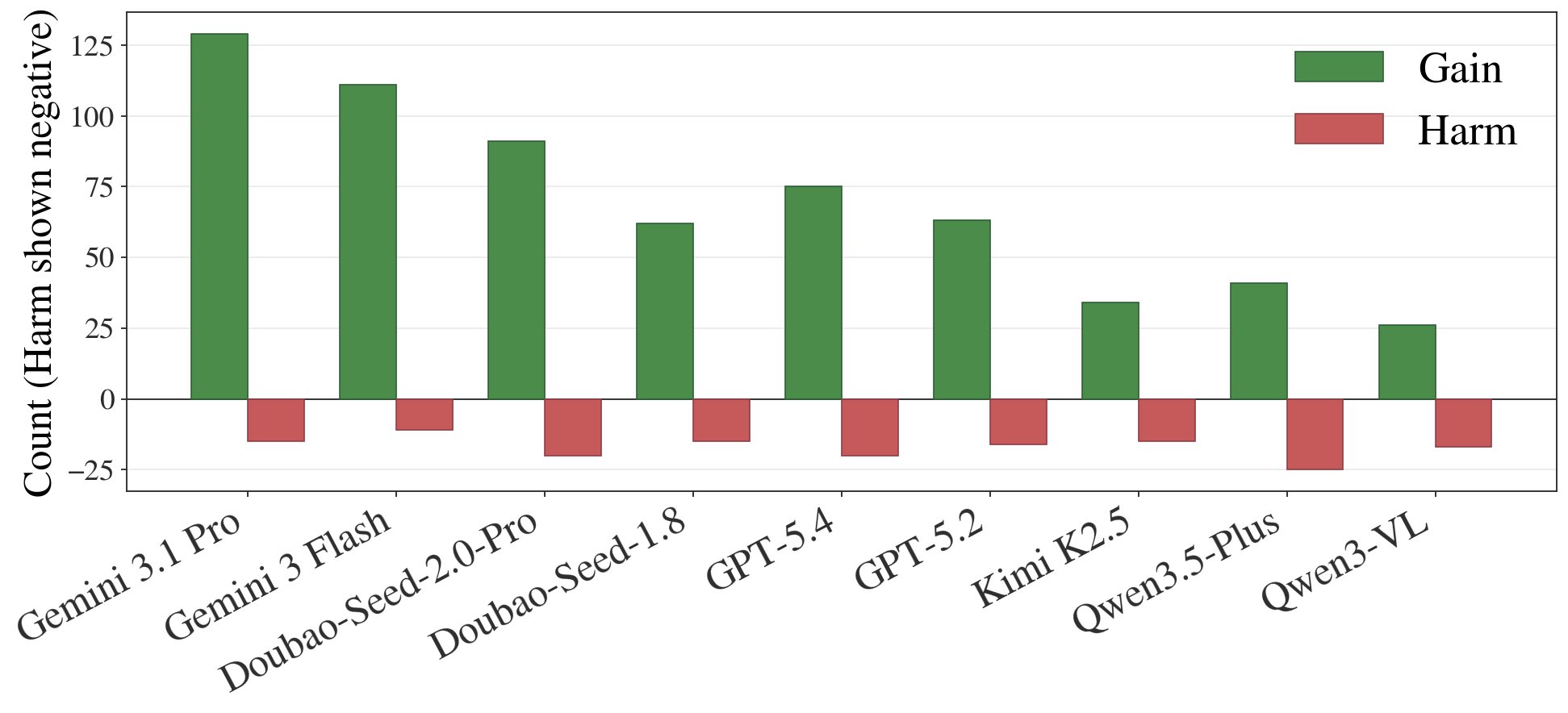}
  \end{minipage}
  \hfill
  \begin{minipage}[t]{0.48\textwidth}
    \centering
    \includegraphics[width=\textwidth]{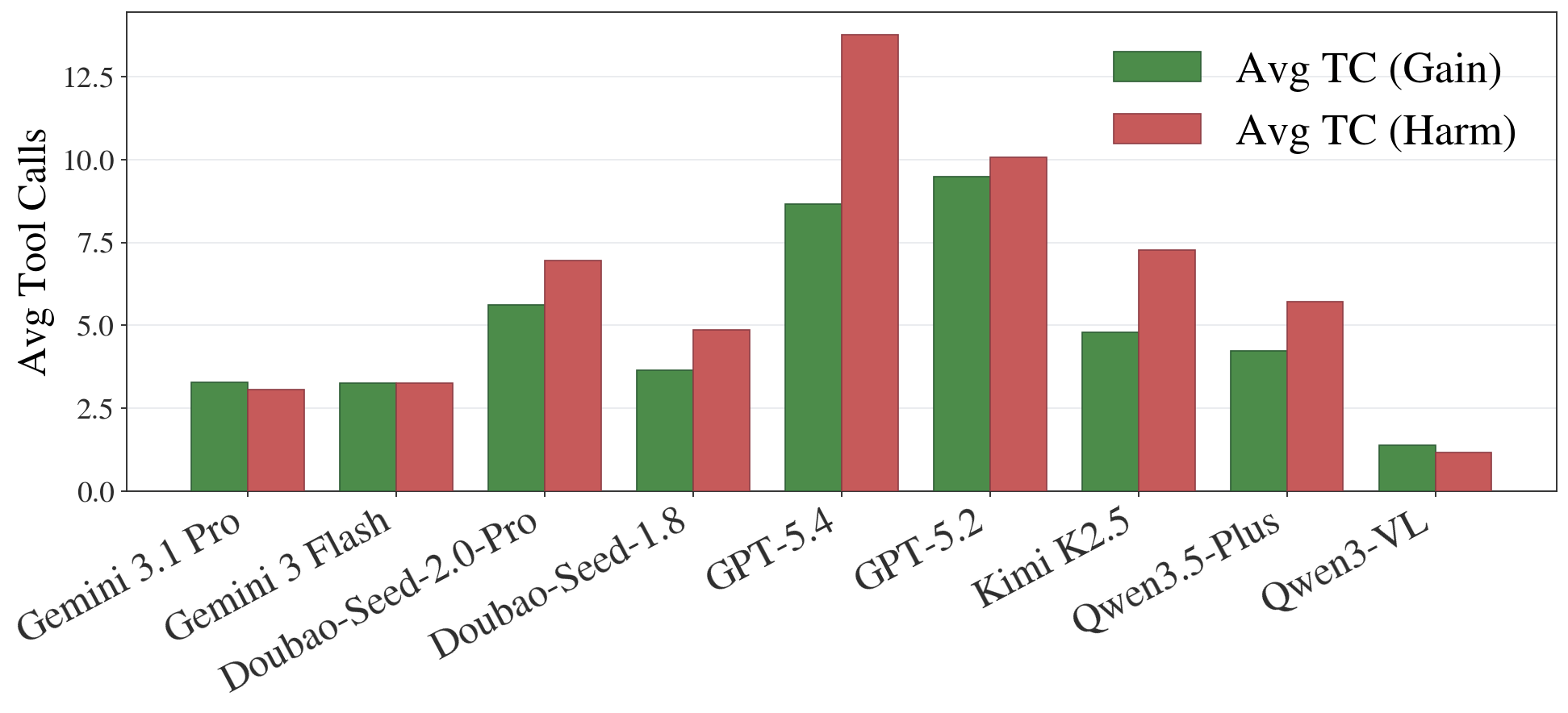}
  \end{minipage}
  \caption{
  Stability and efficiency analysis of tool use. The left panel shows
  Gain/Harm analysis: Gain denotes questions answered incorrectly without
  tools but correctly with tools, and Harm denotes questions answered
  correctly without tools but incorrectly with tools. The right panel shows
  the average number of tool calls on Gain and Harm samples.
  }
  \label{fig:tool_efficiency}
\end{figure*}

Beyond final accuracy, we analyze tool-use behavior to understand why some
models can call tools but still fail to answer correctly. This analysis is
not a separate benchmark objective; instead, it provides behavioral
evidence about search reliability and error amplification under the
with-tools protocol. We adopt the Gain/Harm analysis
framework~\citep{ma2026med}, counting for each model the number of
questions that become correct with tools compared with the setting without tools
(Gain) and the number of questions that become wrong (Harm). We also
compute the average number of tool calls on Gain and Harm samples to
characterize the interaction cost behind positive and negative changes.

Figure~\ref{fig:tool_efficiency} shows that Gemini 3.1 Pro and Gemini 3
Flash have the most stable positive effects: their Gain counts are much
higher than their Harm counts (129 vs.\ 15; 111 vs.\ 11), and their average
numbers of calls on both Gain and Harm samples remain low. These models can
therefore correct original errors with tools while usually completing
effective search at low interaction cost.

By contrast, although GPT-5.4 and GPT-5.2 also benefit from tools to some
extent, their gains come with substantially higher instability. They still
have many Gain samples (75 and 63), but their Harm counts also rise to 20
and 16, and their average numbers of calls on Harm samples are higher. This
suggests that their problem is not insufficient calling frequency; rather,
their long interaction processes still struggle to converge reliably to the
correct evidence. Weaker models further show that simply using tools is
insufficient for improvement. For example, Qwen3-VL-235B-A22B-Thinking has
only limited net benefit (26 Gain vs.\ 17 Harm), indicating that tool use
itself can amplify errors when a model cannot use tools reliably.

\subsection{Search Efficiency Analysis}

\begin{figure*}[t]
  \centering
  \begin{minipage}[t]{0.48\textwidth}
    \centering
    \includegraphics[width=\textwidth]{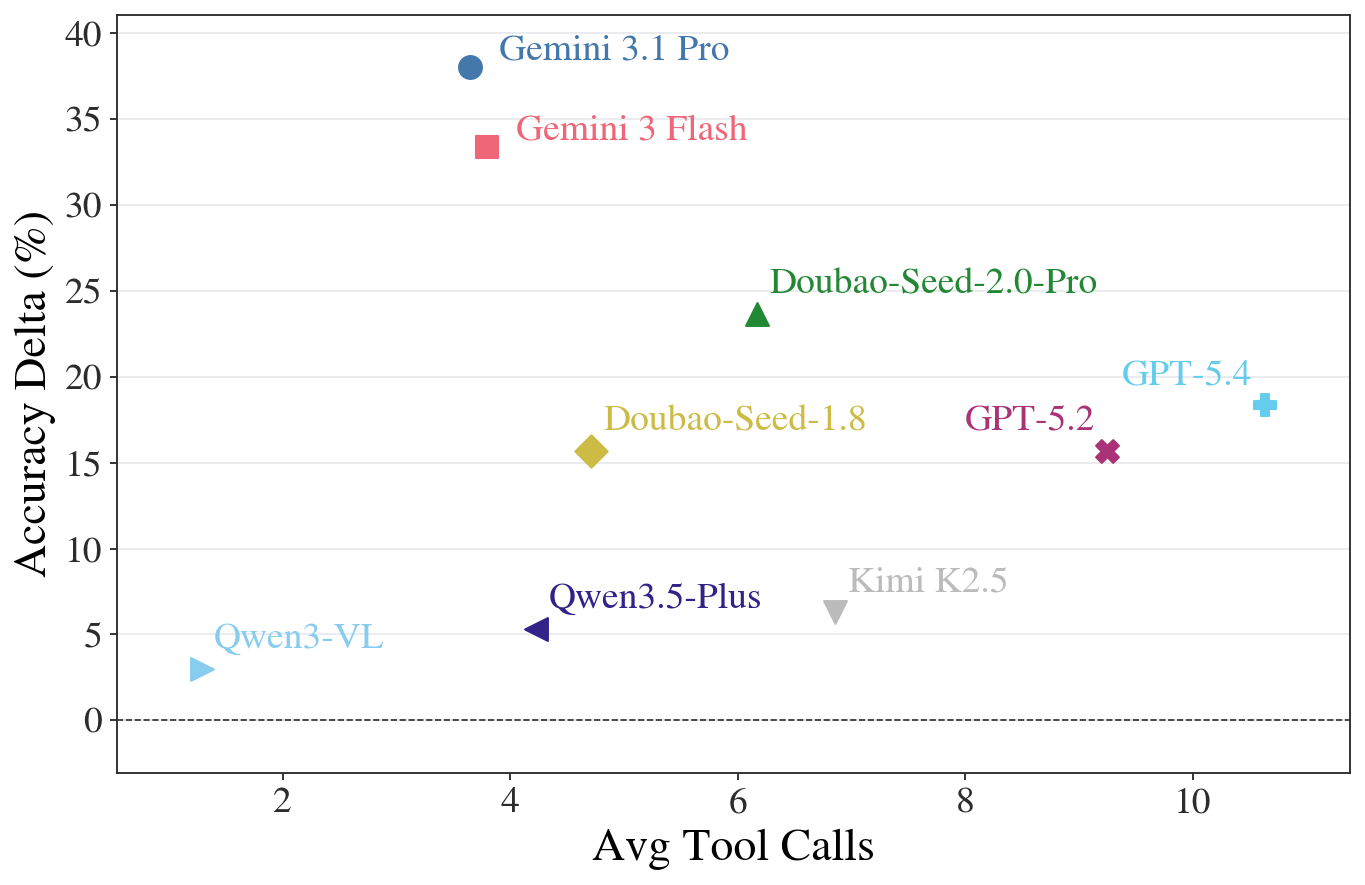}
  \end{minipage}
  \hfill
  \begin{minipage}[t]{0.48\textwidth}
    \centering
    \includegraphics[width=\textwidth]{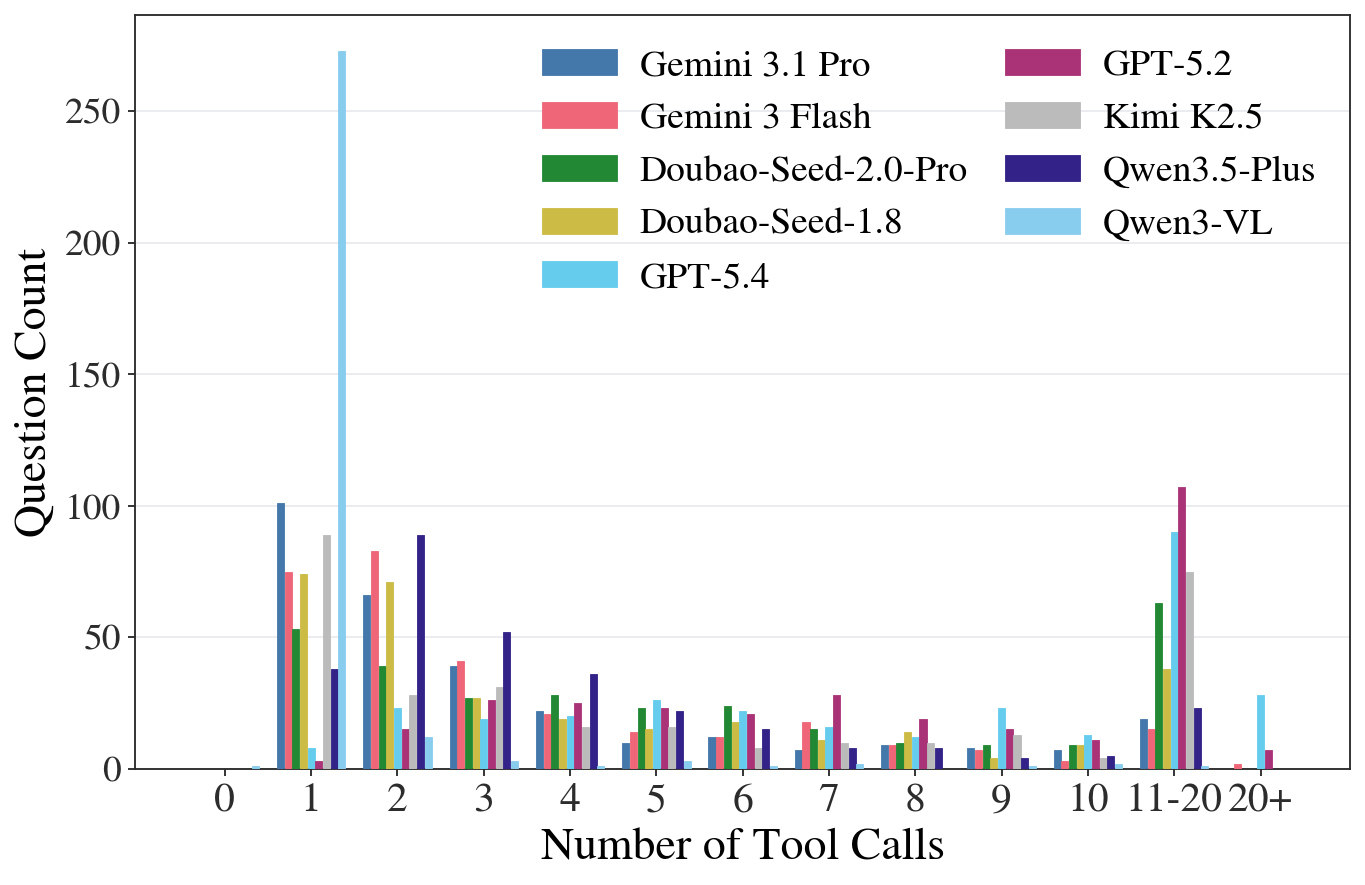}
  \end{minipage}
  \caption{
  Analysis of models' tool-use efficiency and calling behavior. The left
  panel shows the relationship between average tool calls and accuracy gain
  among samples with tool calls. The right panel shows the frequency
  distribution of questions across different tool-call intervals for each
  model, characterizing their tool-use behavior patterns.
  }
  \label{fig:search_efficiency}
\end{figure*}

We further characterize tool-use behavior through call intensity and call
distribution. This analysis helps explain the gap between tool availability
and effective evidence use: models may call Crop many times without
converging to the right local evidence, or stop early before entering the
relevant region. As shown in the left panel of
Figure~\ref{fig:search_efficiency}, Gemini 3.1 Pro and Gemini 3 Flash are
closer to the upper-left region, indicating that they obtain higher
accuracy gains with fewer tool calls. This is consistent with their low average
tool-call counts in the main table. By contrast, GPT-5.4 and GPT-5.2 have
higher call intensity, but their gains do not increase correspondingly,
showing that more frequent interaction does not automatically translate
into more effective search. Meanwhile, Qwen3-VL-235B-A22B-Thinking has low
average call counts, but its gain remains small, suggesting that low call
frequency can also reflect premature stopping or ineffective localization.

The same difference is also visible in the call distribution shown in the
right panel of Figure~\ref{fig:search_efficiency}. Calls from Gemini 3.1 Pro
and Gemini 3 Flash are mainly concentrated in the 1--3 range, suggesting that
these models usually complete effective localization early. By contrast,
GPT-5.4, GPT-5.2, and Kimi K2.5 have higher proportions in the 11--20
range, showing more pronounced long-tail calling behavior. These behavioral
patterns explain why accuracy with tools differs across models even under
the same Crop interface: successful active visual search requires not only
calling tools, but also stopping at the right local evidence and integrating
it into the final answer.

A trajectory-level analysis of target coverage and search steps is
provided in Appendix~\ref{app:crop_coverage}. It distinguishes the steps
needed to first cover the annotated target from subsequent Crop actions.

\section{Conclusion}

We introduced \textbf{VisualNeedle}, a 300-question
``needle-in-a-haystack'' benchmark whose construction closes three
shortcuts that let MLLMs score high without genuinely engaging with
the image: language priors, coarse global semantics, and
near-invariance to the content of intermediate tool outputs.
Across 9 mainstream models, accuracy without tools stays below 20\%;
in the text-only diagnostic, accuracy stays below 10\%. The human majority
baseline is 63.00\%, while the strongest evaluated model, Gemini 3.1 Pro,
reaches only 56.00\% on average even with visual tools. The crop-black setting further shows that real crops improve all evaluated
models by 3.77--44.00 percentage points over black crops on
VisualNeedle. By contrast, all models in the V* Bench diagnostic still reach
75.92\%--87.43\% under the crop-black setting, and several strong
HR-Bench results remain at or above 82.00\%, showing that prior
benchmarks can be largely solved with limited dependence on
intermediate visual content. VisualNeedle thus provides a diagnostic,
shortcut-resistant testbed for active visual search and clearer
evaluation targets for future visual agent systems.

\section*{AI use statement}

We used generative AI tools to aid and polish the writing of this paper,
including improving the clarity, grammar, and organization of the text.

\section*{Reproducibility Statement}

The VisualNeedle dataset and evaluation code are provided in the
supplementary material. Appendix~\ref{app:implementation} describes the
dataset format, image preprocessing, evaluation settings, generation
hyperparameters, termination criteria, and answer-scoring procedure.
Appendix~\ref{app:prompts} gives the model and judge prompts and the
Crop tool interface. Appendices~\ref{app:human_annotation}
and~\ref{app:judge_audit} describe the human evaluation protocol and the
judge audit, respectively. We repeat the main evaluations without and with
tools three times and report the means, standard deviations, and
confidence intervals in Appendix~\ref{app:repeat_variance}. Because
several evaluated models are accessed through closed-source APIs, exact
numerical results may vary due to sampling stochasticity and later
changes to the underlying services.

\section*{Ethics Statement}
\label{sec:ethics}

VisualNeedle is constructed for research on fine-grained visual
understanding and active visual search. The images were obtained through
a compliant data procurement process rather than unauthorized web
scraping, and the data provider was responsible for securing the
permissions required for data collection, use, and public release.
To reduce privacy risks, we mask human faces and vehicle license plates
in the dataset images, and we exclude pornographic, violent, and
politically sensitive content. We use the dataset solely for
research-oriented model evaluation and analysis, and not for identifying
or tracking individuals or for commercial purposes.

\bibliographystyle{iclr2027_conference}
\bibliography{iclr2027_conference}

\clearpage
\appendix
\raggedbottom

\section{Limitations}

VisualNeedle is designed as a diagnostic benchmark for active visual search in
information-dense scenes, but it has several limitations. First, the
benchmark contains 300 carefully curated questions, which enables detailed
annotation and trajectory analysis but cannot exhaustively cover all real-world
visual search scenarios, languages, cultural contexts, or image sources.
Second, several evaluated systems are closed-source API models, so their
results may be affected by model-version updates, service-side changes, and
sampling stochasticity. Third, because VisualNeedle uses open-ended questions,
normalized exact matching alone is insufficient for answer evaluation;
therefore, we use a separate VLM judge (Gemini 3.1 Pro) after exact matching, which
improves coverage but may still introduce small errors for ambiguous visual
evidence, equivalent answer expressions, or borderline cases.

\section{Implementation Details}
\label{app:implementation}

In implementation, we use a unified agent evaluation framework for all experiments. In the with-tools setting, a model receives the full image as its initial input and can call a visual crop tool over multiple turns to progressively zoom into local regions. In the without-tools setting, tool-calling ability is removed, and the model must answer directly from the full image. To avoid state contamination across samples, we instantiate a fresh agent for each sample and assign it an independent tool working directory. Input images are converted to local files before evaluation and compressed to no more than 36M pixels when necessary. During evaluation, we record the parameters and outputs of each tool call as well as derived images, preserving complete search trajectories. Answer judgment follows a two-stage strategy: we first perform normalized exact matching; if rule-based matching fails, we then call a separate VLM judge (Gemini 3.1 Pro), which determines whether the model output is semantically correct given the original image, question, ground-truth answer, and model response.

\paragraph{Sample format.}
VisualNeedle is stored as a JSONL dataset with one question per line. Each
sample contains a unique annotation identifier \texttt{id}, an image identifier
\texttt{tag}, a local image path \texttt{image}, a hosted image URL
\texttt{image\_url}, the question \texttt{question}, the ground-truth answer
\texttt{answer}, category labels \texttt{class1}, \texttt{class2}, and
\texttt{class3}, a formatting flag \texttt{is\_markdown}, and a ground-truth
target box \texttt{bbox}. The released evaluation split used in this paper does
not include a separate alternative-answer field; equivalent answer variants are
handled by the answer normalizer and the Gemini 3.1 Pro VLM judge described
below. The target box is stored in the original image coordinate frame, not in
the resized model-input frame. Its format is \texttt{[x1, y1, x2, y2]}, where
\texttt{(x1, y1)} is the top-left corner and \texttt{(x2, y2)} is the
bottom-right corner. We use the standard half-open image-box convention for
area, overlap, and cropping computations: valid boxes satisfy
\texttt{0 <= x1 < x2 <= W} and \texttt{0 <= y1 < y2 <= H}, so the right and
bottom boundaries are treated as exclusive.

\paragraph{Evaluation settings.}
All models are evaluated in a zero-shot setting under four protocols. In
\textit{text-only}, the model receives only the question, with both image fields
removed. In the \textit{without tools} (w/o tools) setting, the model receives the full image but the tool
interface is disabled, so it must answer from the initial global view. In
the \textit{with tools} (w/ tools) setting, the model receives the full image and may call the Crop tool
over multiple turns; each Crop call specifies an \texttt{img\_idx}, a
\texttt{bbox\_2d}, and a free-form \texttt{reason}, and returns the selected
high-resolution local crop as the next visual observation. In the
crop-black setting, the interaction and tool schema are identical to
the with-tools setting, but every returned crop image is replaced in place by an
all-black image with exactly the same pixel size as the true crop. Thus this
intervention preserves the interaction protocol, tool interface, and output
dimensions while removing the visual evidence inside the crop; the model
continues to decide when and where to call the tool on its own.

\paragraph{Termination and scoring.}
All models are connected to the same evaluation framework through a unified API
adaptation layer. For each question, the model may make at most 100 tool calls.
The interaction ends when the model emits a final answer without a pending tool
call or exhausts this budget. If the model does not produce a parsable answer,
exceeds the per-sample timeout, repeatedly produces malformed or illegal tool
calls that cannot be recovered by the evaluation harness, or terminates without
an answer, the sample is counted as incorrect. For answer judgment, we first
apply normalized exact matching. If exact matching fails, we use a separate VLM
judge (Gemini 3.1 Pro), which receives the original image, the question, the
ground-truth answer, the parsed answer, and the full model response. The judge
returns a binary correctness decision; only a positive decision changes an
exact-match failure into a correct prediction, and invalid or negative judge
outputs are counted as incorrect.

The temperature and thinking settings reported for each model are
summarized in Table~\ref{tab:eval_hparams}. We use
\texttt{thinking\_level} as a common reporting label for the models'
reasoning or thinking settings; it does not imply that all model APIs
use this parameter name.

\begin{table}[th]
  \centering
  \caption{Generation hyperparameters used in evaluation.}
  \label{tab:eval_hparams}
  \footnotesize
  \setlength{\tabcolsep}{6pt}
  \begin{tabular}{>{\raggedright\arraybackslash}p{0.30\textwidth}>{\raggedright\arraybackslash}p{0.60\textwidth}}
    \toprule
    Model(s) & Hyperparameters \\
    \midrule
    GPT-5.2, GPT-5.4 &
    \texttt{temperature = 1.0}, \texttt{thinking\_level = xhigh}. \\
    Gemini 3.1 Pro, Gemini 3 Flash &
    \texttt{temperature = 1.0}, \texttt{thinking\_level = high},
    \texttt{max\_tokens = 65,536}. \\
    Doubao-Seed-1.8 &
    \texttt{temperature = 1.0}, \texttt{thinking\_level = high},
    \texttt{max\_tokens = 32,000}. \\
    Doubao-Seed-2.0-Pro &
    \texttt{temperature = 1.0}, \texttt{thinking\_level = high},
    \texttt{max\_tokens = 128,000}. \\
    Kimi K2.5 &
    \texttt{temperature = 1.0}, \texttt{thinking\_enabled},
    \texttt{max\_tokens = 128,000}. \\
    Qwen3.5-Plus &
    \texttt{temperature = 1.0}, \texttt{thinking\_enabled}. \\
    Qwen3-VL-235B-A22B-Thinking &
    \texttt{temperature = 0.7}, \texttt{top\_p = 0.8},
    \texttt{top\_k = 20}, \texttt{repetition\_penalty = 1.0},
    \texttt{presence\_penalty = 1.5}. \\
    \bottomrule
  \end{tabular}
\end{table}

\section{Target-Region Size Distribution}
\label{app:target_region_size}

To characterize the spatial extent of the decisive visual evidence, we
measure each annotated target box as a percentage of the original full
image area. For an original image of width $W$ and height $H$, and a
ground-truth target box $[x_1,y_1,x_2,y_2]$, the relative area is
$100(x_2-x_1)(y_2-y_1)/(WH)\%$. We use the original image dimensions and
annotation coordinates described in the sample format above, rather than
the dimensions of a resized model input or a returned crop.

\begin{table}[t]
  \centering
  \caption{Distribution of the ground-truth target-box area relative to the
  original full-image area across the 300 VisualNeedle questions.}
  \label{tab:target_region_size}
  \footnotesize
  \setlength{\tabcolsep}{5pt}
  \begin{tabular}{lrr}
    \toprule
    Target-box area / Full image & Samples & Proportion \\
    \midrule
    $<0.01\%$       & 0   & 0.00\% \\
    $0.01$--$0.03\%$ & 41  & 13.67\% \\
    $0.03$--$0.10\%$ & 109 & 36.33\% \\
    $0.10$--$0.30\%$ & 114 & 38.00\% \\
    $0.30$--$1.00\%$ & 33  & 11.00\% \\
    $1.00$--$3.00\%$ & 2   & 0.67\% \\
    $>3.00\%$        & 1   & 0.33\% \\
    \midrule
    \textbf{Total} & \textbf{300} & \textbf{100.00\%} \\
    \bottomrule
  \end{tabular}
\end{table}

Table~\ref{tab:target_region_size} shows that 297 of the 300 target
boxes (99.00\%) occupy no more than 1.00\% of the original image area.
The highest-count intervals are $0.10$--$0.30\%$ (114 questions) and
$0.03$--$0.10\%$ (109 questions). Thus, the evidence required to answer
most questions occupies only a small fraction of the global scene. This
area statistic describes the annotated targets; whether a model actually
finds and uses them is assessed separately through crop trajectories in
Appendix~\ref{app:crop_coverage}.

\section{Human Annotation Protocol}
\label{app:human_annotation}

To ensure a rigorous and unbiased assessment, we recruited an independent evaluation team consisting of 12 professional annotators. Each team member possesses extensive experience in multimodal data labeling and reasoning. Importantly, these annotators were held blind to all preceding stages of the pipeline, including image collection, question authoring, bounding-box annotation, and quality filtering. This separates the human-baseline answering
stage from the data-construction and verification stage described in
Section~\ref{sec:curation_pipeline}. Each evaluation item was answered independently by 3 different
annotators, yielding 900 human annotations in total. Annotators saw one
image--question pair at a time and submitted a single best answer in the same
answer format expected from the models. They were not shown model-generated
hints, answers from other annotators, ground-truth answers, target boxes, or
other construction-time metadata.

All annotators used the same web-based annotation platform used for human
evaluation, with the same user interface, image viewer, and shortcut bindings.
The image viewer
allowed unrestricted zooming and panning over the original-resolution image, so
annotators could inspect arbitrary local regions before answering. Human
annotators did not call the model-side Crop API; their interface provided the
human-side analogue of free local visual search through zoom and pan. No
annotator used a custom client or external image editor. For fairness,
annotators were instructed to spend at most 5 minutes per item; if they could
not locate the answer with reasonable confidence after this budget, they were
asked to submit their best guess or mark the item as suspended. In the final
300-item set, all assigned items received actual answers, giving a 100.00\%
completion rate and a 0.00\% suspension rate. Platform logs record the assigned
annotator, start/end timestamps, elapsed time, submitted answer, and suspension
flag for each item.

\begin{table}[th]
  \centering
  \caption{Human annotation protocol for the human-baseline evaluation on the final 300-item split.}
  \label{tab:human_annotation_protocol}
  \footnotesize
  \setlength{\tabcolsep}{4pt}
  \begin{tabular}{p{0.32\textwidth}p{0.62\textwidth}}
    \toprule
    Aspect & Protocol \\
    \midrule
    Annotator pool & 12 external annotators; no overlap with the data-construction team. \\
    Annotations per item & 3 independent annotators per item, yielding 900 annotations in total. \\
    Task input & One high-resolution image and its natural-language question. \\
    Task output & One answer string per item; no rationale required. \\
    Hidden information & No model hints, other annotators' answers, ground-truth answers, target boxes, difficulty tags, or filtering metadata. \\
    Zoom / crop access & Unrestricted zoom and pan in the platform image viewer; no model-side Crop API, external image editor, or custom client. \\
    Shared human interface & All annotators used the same web-based human-evaluation platform, UI, viewer, and shortcuts. \\
    Time budget & At most 5 minutes per item; submit best guess or suspend if unresolved. \\
    Measured time & Mean per-annotator time ranged from 2 min 37 s to 3 min 35 s; overall mean was approximately 3 min 17 s per item. \\
    Completion / suspension & 100.00\% completion rate and 0.00\% suspension rate on the final 300-item set. \\
    Reported metrics & Majority Vote@3: 189/300 = 63.00\%; Pass@3: 224/300 = 74.67\%. \\
    \bottomrule
  \end{tabular}
\end{table}

\section{Judge Audit and Case Analysis}
\label{app:judge_audit}

All evaluation samples come with reference answers (GT), and the data
construction process aims to make each answer as unique and unambiguous as
possible. For answer scoring, we first apply normalized exact matching. If
exact matching fails, Gemini 3.1 Pro performs a VLM-based semantic check using
the image, question, model prediction, and reference answer. In other words,
the judge combines semantic matching with reference-answer analysis rather
than relying on string equality alone.

\subsection{Human Audit of Automatic Grading}

The reported accuracy of open-ended visual answers depends partly on the reliability of our automatic grading procedure. Exact matching handles unambiguous string matches, whereas the Gemini 3.1 Pro judge assesses semantic equivalence when exact matching fails. To examine whether this procedure agrees with human judgment, and whether using an evaluated model as the semantic judge might favor its own answers, three human quality-control reviewers audited the responses from six 300-question model runs (1,800 responses in total). We compared the final binary decision of the automatic grading procedure with the human audit for each response and recomputed accuracy for each audited run. Four of the six audited runs are the first with-tools runs of their corresponding models in the main experiment. A fifth run evaluates Gemini 3 Pro, which is not included in the main comparison; the sixth is an additional, independent Gemini 3.1 Pro run outside the three main-experiment runs, included to enlarge the check on the judge model’s own answers. These six runs constitute the audit set. Their individual accuracies should not be conflated with the three-run means in the main results table; neither the Gemini 3 Pro run nor the additional Gemini 3.1 Pro run contributes to those means.

\begin{table}[t]
  \centering
  \caption{Agreement between automatic grading and human audit across
  1,800 responses from six model runs. Agreement comprises the two
  diagonal cells. Cohen's $\kappa$ adjusts observed agreement for
  chance agreement implied by the two sets of label frequencies.}
  \label{tab:judge_human_agreement}
  \footnotesize
  \setlength{\tabcolsep}{5pt}
  \begin{tabular*}{\columnwidth}{@{\extracolsep{\fill}}llr@{}}
    \toprule
    Automatic grading & Human audit & Responses \\
    \midrule
    Correct & Correct & 790 \\
    Incorrect & Incorrect & 946 \\
    Correct & Incorrect & 61 \\
    Incorrect & Correct & 3 \\
    \midrule
    \multicolumn{2}{l}{Overall agreement} & 1,736/1,800 (96.44\%) \\
    \multicolumn{2}{l}{Cohen's $\kappa$} & 0.9284 \\
    \bottomrule
  \end{tabular*}
\end{table}

As shown in Table~\ref{tab:judge_human_agreement}, automatic grading
and human audit agreed on 1,736 of the 1,800 responses (96.44\%),
including 790 judged correct by both and 946 judged incorrect by both.
The 64 disagreements were asymmetric: the automatic procedure accepted
61 responses that the human audit marked incorrect, whereas it rejected
only 3 responses that the human audit marked correct. Cohen's
$\kappa=0.9284$ indicates high agreement after accounting for chance
agreement under the observed label frequencies. Automatic grading
marked 851/1,800 responses correct (47.28\%), compared with 793/1,800
under human audit (44.06\%), a difference of 3.22 percentage points.
Thus, despite high overall agreement, the automatic procedure tended
to be more permissive on the audited disagreements.

\begin{table}[t]
  \centering
  \caption{Human audit of six 300-question model runs. Accuracy change
  is human-audited accuracy minus automatic-grading accuracy, in
  percentage points (pp). Agreement is the percentage of responses for
  which the two methods assign the same correctness label. Displayed
  accuracy changes are calculated from unrounded values.}
  \label{tab:judge_human_by_run}
  \footnotesize
  \setlength{\tabcolsep}{4pt}
  \begin{tabular}{lrrrr}
    \toprule
    Evaluated model
    & \shortstack{Automatic-grading\\Accuracy (\%)}
    & \shortstack{Human-audited\\Accuracy (\%)}
    & \shortstack{Accuracy\\Change (pp)}
    & \shortstack{Automatic--human\\Agreement (\%)} \\
    \midrule
    Doubao-Seed-1.8 & 33.33 & 28.67 & $-4.67$ & 95.33 \\
    Doubao-Seed-2.0-Pro & 38.00 & 34.00 & $-4.00$ & 96.00 \\
    Gemini 3 Flash & 47.00 & 43.00 & $-4.00$ & 96.00 \\
    Gemini 3 Pro & 57.00 & 53.33 & $-3.67$ & 96.33 \\
    Gemini 3.1 Pro (main run) & 55.33 & 52.33 & $-3.00$ & 97.00 \\
    Gemini 3.1 Pro (additional run) & 53.00 & 53.00 & $0.00$ & 98.00 \\
    \bottomrule
  \end{tabular}
\end{table}

Table~\ref{tab:judge_human_by_run} shows that automatic--human
agreement exceeded 95\% in every audited run. Human audit reduced
the reported accuracy for five runs and left one unchanged. The
relative ordering of models was largely stable: Gemini 3 Pro and the
two Gemini 3.1 Pro runs remained at the top of this audit set, while
the two Doubao runs remained below Gemini 3 Flash. The ordering of
the two Gemini 3.1 Pro runs did change, so the audit does not imply
that every pairwise rank was preserved.

Because Gemini 3.1 Pro also served as the semantic judge, its two
audited runs provide a specific check for possible self-favoring in
this audit set. Their automatic--human agreement rates were 97.00\%
and 98.00\%, at least as high as those for the other four runs.
Human audit decreased the main run's accuracy by 3.00 points and did
not change the additional run's accuracy. We therefore find no clear
evidence of systematic self-favoring in these two audited runs; this
observation does not rule out bias outside the audited responses or
other settings.

\subsection{Illustrative Judge Decisions}

Representative cases are shown below.

\begin{casebox}{Good case: hard but judged correctly}
\textbf{Question.} What are the characters on the sign to the right of
``\zh{裘记单车修理铺}''?

\textbf{Prediction.} \zh{诚兴麻雀公司}

\textbf{Gold.} \zht{誠興麻雀公司}
\end{casebox}

\begin{casebox}{Good case: hard translation equivalent}
\textbf{Question.} What is the name of the sixth book on the shelf below
``\zh{朝花夕拾}''?

\textbf{Prediction.} \zh{百年孤独}

\textbf{Gold.} One Hundred Years of Solitude
\end{casebox}

\begin{casebox}{Bad case: false negative}
\textbf{Question.} In the image, what characters are written on the sycee held
by the lucky cat graffiti to the left of the man wearing a hat?

\textbf{Prediction.} \zht{開運招財}

\textbf{Gold.} \zht{開運}
\end{casebox}

\begin{casebox}{Bad case: false positive}
\textbf{Question.} What is the second item to the left of the monkey in the
picture?

\textbf{Prediction.} A Hello Kitty doll

\textbf{Gold.} Cat
\end{casebox}

\section{Repeated-run Stability Analysis}
\label{app:repeat_variance}

To assess run-to-run stability, we repeat the main evaluations without and
with tools three times under the same experimental setting.
For each metric $x$, Table~\ref{tab:repeat_accuracy_variance} and
Table~\ref{tab:repeat_call_variance} report the mean, sample standard
deviation, and 95\% confidence interval half-width:
\[
\bar{x}=\frac{1}{3}\sum_{i=1}^{3}x_i,\quad
s=\sqrt{\frac{1}{2}\sum_{i=1}^{3}(x_i-\bar{x})^2},\quad
\mathrm{CI}_{95}=t_{0.975,2}\frac{s}{\sqrt{3}},
\]
where $t_{0.975,2}=4.303$. These intervals capture stochastic variation
across repeated evaluations on the fixed 300-question split, rather than
uncertainty from resampling the benchmark itself. Because only three
repetitions are available, the confidence intervals should be interpreted
as a conservative stability diagnostic rather than as a definitive
population-level estimate.

\begin{table}[!htb]
  \centering
  \caption{
  Run-to-run variance for accuracy metrics.
  Each accuracy entry reports the mean accuracy in percent and the sample
  standard deviation in percentage points across three repeated runs.
  $\Delta$ denotes the absolute accuracy improvement in percentage points.
  CI$_{95}$ reports the 95\% confidence-interval half-width in percentage
  points.
  }
  \label{tab:repeat_accuracy_variance}
  \footnotesize
  \setlength{\tabcolsep}{3pt}
  \begin{tabular}{lcccccc}
    \toprule
    Model
    & \shortstack{w/ tools\\Mean (\%)\\$\pm$ Std. (pp)}
    & \shortstack{w/ tools\\CI$_{95}$\\(pp)}
    & \shortstack{w/o tools\\Mean (\%)\\$\pm$ Std. (pp)}
    & \shortstack{w/o tools\\CI$_{95}$\\(pp)}
    & \shortstack{$\Delta$\\Mean\\$\pm$ Std. (pp)}
    & \shortstack{$\Delta$\\CI$_{95}$\\(pp)} \\
    \midrule
    Gemini 3.1 Pro
    & 56.00 $\pm$ 0.67
    & 1.66
    & 16.33 $\pm$ 0.88
    & 2.19
    & 39.67 $\pm$ 1.53
    & 3.79 \\

    Gemini 3 Flash
    & 48.00 $\pm$ 1.45
    & 3.61
    & 15.11 $\pm$ 1.71
    & 4.25
    & 32.89 $\pm$ 2.36
    & 5.88 \\

    Doubao-Seed-2.0-Pro
    & 36.78 $\pm$ 1.84
    & 4.56
    & 13.11 $\pm$ 1.17
    & 2.91
    & 23.67 $\pm$ 2.00
    & 4.97 \\

    Doubao-Seed-1.8
    & 31.78 $\pm$ 1.39
    & 3.45
    & 17.89 $\pm$ 1.02
    & 2.53
    & 13.89 $\pm$ 1.68
    & 4.17 \\

    GPT-5.4
    & 31.22 $\pm$ 1.07
    & 2.66
    & 14.33 $\pm$ 0.67
    & 1.66
    & 16.89 $\pm$ 1.71
    & 4.25 \\

    GPT-5.2
    & 28.44 $\pm$ 1.84
    & 4.56
    & 10.56 $\pm$ 1.50
    & 3.73
    & 17.89 $\pm$ 2.34
    & 5.82 \\

    Qwen3.5-Plus
    & 22.11 $\pm$ 0.38
    & 0.96
    & 15.00 $\pm$ 1.76
    & 4.38
    & 7.11 $\pm$ 1.54
    & 3.82 \\

    Kimi K2.5
    & 19.33 $\pm$ 3.18
    & 7.90
    & 10.78 $\pm$ 1.50
    & 3.73
    & 8.56 $\pm$ 2.04
    & 5.06 \\

    Qwen3-VL-235B-A22B-Thinking
    & 11.44 $\pm$ 0.84
    & 2.08
    & 8.33 $\pm$ 0.67
    & 1.66
    & 3.11 $\pm$ 0.19
    & 0.48 \\
    \bottomrule
  \end{tabular}
\end{table}

\begin{table}[th]
  \centering
  \caption{Tool-call stability across repeated runs. Avg. Tool Calls is the
  per-question average number of Crop calls, and Total Tool Calls is the
  total number of Crop calls over all 300 questions.}
  \label{tab:repeat_call_variance}
  \footnotesize
  \setlength{\tabcolsep}{3pt}
  \begin{tabular}{lcccc}
    \toprule
    Model
    & \shortstack{Avg. Tool Calls\\Mean $\pm$ Std.}
    & \shortstack{Avg. Tool Calls\\CI$_{95}$}
    & \shortstack{Total Tool Calls\\Mean $\pm$ Std.}
    & \shortstack{Total Tool Calls\\CI$_{95}$} \\
    \midrule
    Gemini 3.1 Pro
    & 3.72 $\pm$ 0.07 & 0.18
    & 1117.33 $\pm$ 22.12 & 54.96 \\

    Gemini 3 Flash
    & 3.91 $\pm$ 0.11 & 0.26
    & 1174.00 $\pm$ 31.76 & 78.91 \\

    Doubao-Seed-2.0-Pro
    & 6.37 $\pm$ 0.19 & 0.47
    & 1912.33 $\pm$ 56.66 & 140.76 \\

    Doubao-Seed-1.8
    & 4.78 $\pm$ 0.10 & 0.24
    & 1434.67 $\pm$ 29.30 & 72.78 \\

    GPT-5.4
    & 11.07 $\pm$ 0.43 & 1.07
    & 3322.00 $\pm$ 129.05 & 320.60 \\

    GPT-5.2
    & 9.19 $\pm$ 0.06 & 0.15
    & 2756.00 $\pm$ 18.08 & 44.92 \\

    Qwen3.5-Plus
    & 3.49 $\pm$ 0.66 & 1.65
    & 1046.00 $\pm$ 199.16 & 494.78 \\

    Kimi K2.5
    & 6.58 $\pm$ 0.37 & 0.92
    & 1974.33 $\pm$ 111.13 & 276.09 \\

    Qwen3-VL-235B-A22B-Thinking
    & 1.58 $\pm$ 0.26 & 0.65
    & 473.33 $\pm$ 78.45 & 194.90 \\
    \bottomrule
  \end{tabular}
\end{table}

The repeated runs support the main qualitative conclusions. Gemini 3.1 Pro
and Gemini 3 Flash consistently form the leading group among the 9 evaluated
models, while Qwen3-VL-235B-A22B-Thinking shows only a small positive gain
with very low run-to-run variation. In contrast, GPT-5.4 and GPT-5.2 have
consistently high tool-call counts, confirming that more interaction alone is
insufficient for effective active visual search.

\section{Interactive Search Process}

This section provides a more formal description of the interactive search process in VisualNeedle. Specifically, at step $t$, the model chooses an action $a_t = (u_t, p_t)$, where $u_t$ denotes the operation selected at the current step and $p_t$ denotes the parameters associated with that operation. The action space is $\mathcal{A} = \{\textbf{Crop}, \textbf{Answer}\}$. When the model chooses \textbf{Crop}, the parameter $p_t$ is the bounding box $bbox_t$; when it chooses \textbf{Answer}, the parameter $p_t$ is the final answer $y$.

\begin{algorithm}[H]
  \caption{Interactive search process in VisualNeedle.}
  \label{alg:task_definition}
  \footnotesize
  \begin{algorithmic}[1]
    \Require High-resolution image $I$, question $Q$
    \State $\mathcal{V} \leftarrow [I]$, $H_0 \leftarrow (Q,I)$, $k \leftarrow 0$
    \While{$k < 100$}
      \State $(u_k,p_k) \leftarrow \pi(H_k)$
      \If{$u_k = \textbf{Answer}$}
        \State \Return $p_k$
      \ElsIf{$u_k = \textbf{Crop}$}
        \State $C_k \leftarrow \textsc{Crop}(\mathcal{V},p_k)$
        \State $\mathcal{V}.\textsc{Append}(C_k)$
        \State $H_{k+1} \leftarrow H_k \mathbin{\Vert} (u_k,p_k,C_k)$
        \State $k \leftarrow k+1$
      \EndIf
    \EndWhile
    \State \Return $\bot$
  \end{algorithmic}
\end{algorithm}

\section{Crop Coverage, Localization Success, and Search Efficiency}
\label{app:crop_coverage}

We analyze tool-use trajectories to separate reaching the target region
from producing a correct final answer. A run is counted as successfully
localized when at least one valid Crop action returns a region that fully
contains the ground-truth target box. Crop coordinates and the target box
are compared in the original image coordinate frame. Localization success
therefore indicates that the model was shown a crop containing the full
annotated target; accuracy with tools measures whether its final answer
was correct. These are distinct marginal outcomes of the same evaluations.

\begin{table*}[t]
  \centering
  \caption{Target coverage and search steps in the w/ tools setting.
  All values are computed over the same three runs as
  Table~\ref{tab:main_results} (900 runs per model). Gap is localization
  success rate minus final accuracy w/ tools in percentage points (pp).
  One search step is one valid Crop action. The three step metrics are
  averaged only over successfully localized runs; Avg. Tool Calls in
  Table~\ref{tab:main_results} is instead averaged over all runs. Each
  step is counted through the final Crop action, including actions after
  first coverage. Step metrics are reported to two decimals; the displayed
  components and total may therefore differ by 0.01 due to rounding.}
  \label{tab:crop_coverage}
  \footnotesize
  \setlength{\tabcolsep}{4pt}
  \begin{tabular}{lrrrrrr}
    \toprule
    Model
    & \shortstack{w/ tools\\Acc. (\%)}
    & \shortstack{Localization\\Success (\%)}
    & \shortstack{Gap\\(pp)}
    & \shortstack{Avg.\\Search\\Steps}
    & \shortstack{Avg. Steps\\to First\\Coverage}
    & \shortstack{Avg. Post-\\coverage\\Refinement} \\
    \midrule
    Gemini 3.1 Pro & 56.00 & 77.78 & 21.78 & 3.67 & 1.85 & 1.82 \\
    Gemini 3 Flash & 48.00 & 67.22 & 19.22 & 3.79 & 1.71 & 2.07 \\
    Doubao-Seed-2.0-Pro & 36.78 & 44.67 & 7.89 & 5.65 & 2.78 & 2.87 \\
    Doubao-Seed-1.8 & 31.78 & 33.78 & 2.00 & 3.64 & 2.15 & 1.49 \\
    GPT-5.4 & 31.22 & 77.00 & 45.78 & 11.95 & 2.20 & 9.75 \\
    GPT-5.2 & 28.44 & 77.00 & 48.56 & 10.01 & 2.61 & 7.40 \\
    Qwen3.5-Plus & 22.11 & 43.44 & 21.33 & 3.57 & 1.50 & 2.07 \\
    Kimi K2.5 & 19.33 & 51.78 & 32.45 & 7.53 & 2.92 & 4.61 \\
    Qwen3-VL-235B-A22B-Thinking & 11.44 & 20.33 & 8.89 & 1.34 & 1.25 & 0.10 \\
    \bottomrule
  \end{tabular}
\end{table*}

The gap in Table~\ref{tab:crop_coverage} is the localization success
rate minus accuracy with tools, measured in percentage points. Every
model has a numerically higher localization success rate than final
accuracy, but the gap varies substantially. Gemini 3.1 Pro reaches
77.78\% localization success and 56.00\% final accuracy, a gap of
21.78 points; Gemini 3 Flash has a 19.22-point gap. GPT-5.4 and GPT-5.2
both reach 77.00\% localization success, essentially matching Gemini
3.1 Pro, but their final accuracies are only 31.22\% and 28.44\%,
giving gaps of 45.78 and 48.56 points. At the other extreme,
Doubao-Seed-1.8 has a gap of only 2.00 points, because its localization
success rate (33.78\%) is itself low and close to its accuracy. High
target coverage alone is therefore insufficient to explain final answer
performance. Since the gap subtracts two marginal rates, it must not be
interpreted as the fraction of localized runs that were answered
incorrectly.

To distinguish stages of search, steps to first coverage count valid
Crop actions up to and including the first crop that fully contains the
target. Post-coverage refinement steps count subsequent valid Crop
actions, whether or not they provide useful new evidence. Total search
steps are the sum of these two counts. Each of the three step metrics is
averaged over successfully localized runs only; in contrast, average
tool calls in the main results are calculated across the full evaluation
split. Gemini 3.1 Pro first covers the target after 1.85 steps on
average and takes a further 1.82 steps, for 3.67 steps in total. For
Gemini 3 Flash the corresponding values are 1.71, 2.07, and 3.79.
These models tend to reach the annotated region early among runs that
successfully localize it, and stop soon afterwards.

The GPT models show a different trajectory pattern. GPT-5.4 reaches
first coverage in 2.20 steps on average, comparable to Gemini 3.1 Pro's
1.85, but performs another 9.75 Crop actions after first coverage,
bringing its total to 11.95 steps. GPT-5.2 takes 2.61 steps to first
coverage and 7.40 further steps, for 10.01 steps in total. Thus, much
of their search activity on successfully localized runs occurs after
the target has already appeared within a returned crop. In particular,
GPT-5.4's low final accuracy cannot be explained by slow first
coverage: it reaches the target about as quickly as Gemini 3.1 Pro on
its successfully localized runs, yet its overall final accuracy remains
substantially lower than its localization success rate. The aggregate
trajectories do not by themselves establish why those answers are
incorrect.

Other models incur a larger share of their search cost before first
coverage. Doubao-Seed-2.0-Pro needs 2.78 steps on average to first cover
the target in successfully localized runs and Kimi K2.5 needs 2.92, the
highest among the evaluated models; their post-coverage counts are 2.87
and 4.61, respectively. Qwen3-VL-235B-A22B-Thinking has the fewest total
steps among successfully localized runs, 1.34, with almost no
post-coverage refinement (0.10), but its localization success and final
accuracy are only 20.33\% and 11.44\%. A short conditional trajectory is
therefore not, on its own, evidence of effective search; it must be read
together with the success rate and answer accuracy.

Taken together, these results distinguish two challenges in active
visual search: obtaining a crop that contains the target and converting
the available local evidence into a correct answer. Gemini 3.1 Pro
combines high localization success, the highest final accuracy, and
few steps on successfully localized runs. GPT-5.4 and GPT-5.2 reach
comparable target coverage but issue many subsequent Crop actions and
end with much lower final accuracy. These trajectory-level measurements
complement the call-frequency analysis in the main paper by showing
where the recorded search steps occur relative to first target coverage.

\section{Category-wise Results}

Table~\ref{tab:category_results} reveals task-difficulty differences in VisualNeedle from a category perspective. In terms of the upper bound of the best model, OCR recognition and spatial relationship are relatively easier, with best accuracies of 67.61\% and 67.24\%, respectively. By contrast, the best accuracies for color recognition and occluded object recognition are below 50\%, and entity recognition barely exceeds it, indicating that current models remain clearly limited when tasks require more fine-grained target confirmation or more complex local reasoning. From the average performance of the 9 models, occluded object recognition is the hardest category, with an average accuracy with tools of only 23.42\%, the lowest among all categories. Its average accuracy without tools is also only 9.91\%, suggesting that even with the Crop tool, models still struggle to reliably integrate local clues for occluded targets, making occlusion reasoning a key weakness of current visual agents. At the same time, OCR recognition has the highest average tool gain, reaching 26.55 percentage points, showing that tools are most effective at alleviating the problem of unclear local details. The average gain for color recognition is only 11.97 percentage points, indicating that in some tasks the bottleneck is no longer only search and zooming, but also subtle attribute discrimination itself.

\begin{figure}[th]
  \centering
  \footnotesize
  \begin{minipage}[t]{0.34\textwidth}
    \centering
    {\footnotesize\textbf{Category Distribution}\par}
    \vspace{2pt}
    \includegraphics[width=\textwidth]{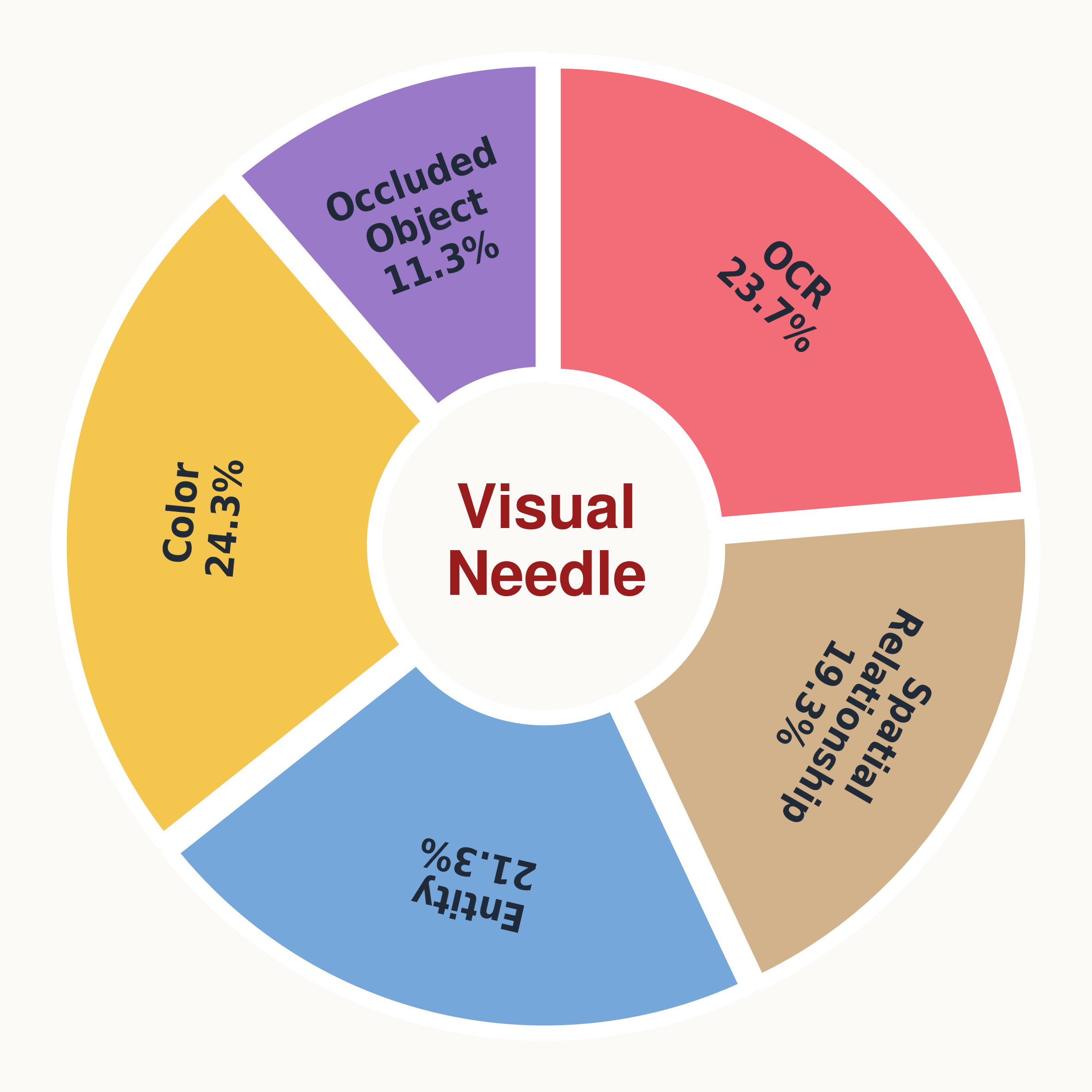}
  \end{minipage}
  \hfill
  \begin{minipage}[t]{0.62\textwidth}
    \centering
    {\footnotesize\textbf{Category Performance}\par}
    \vspace{2pt}
    \includegraphics[width=\textwidth]{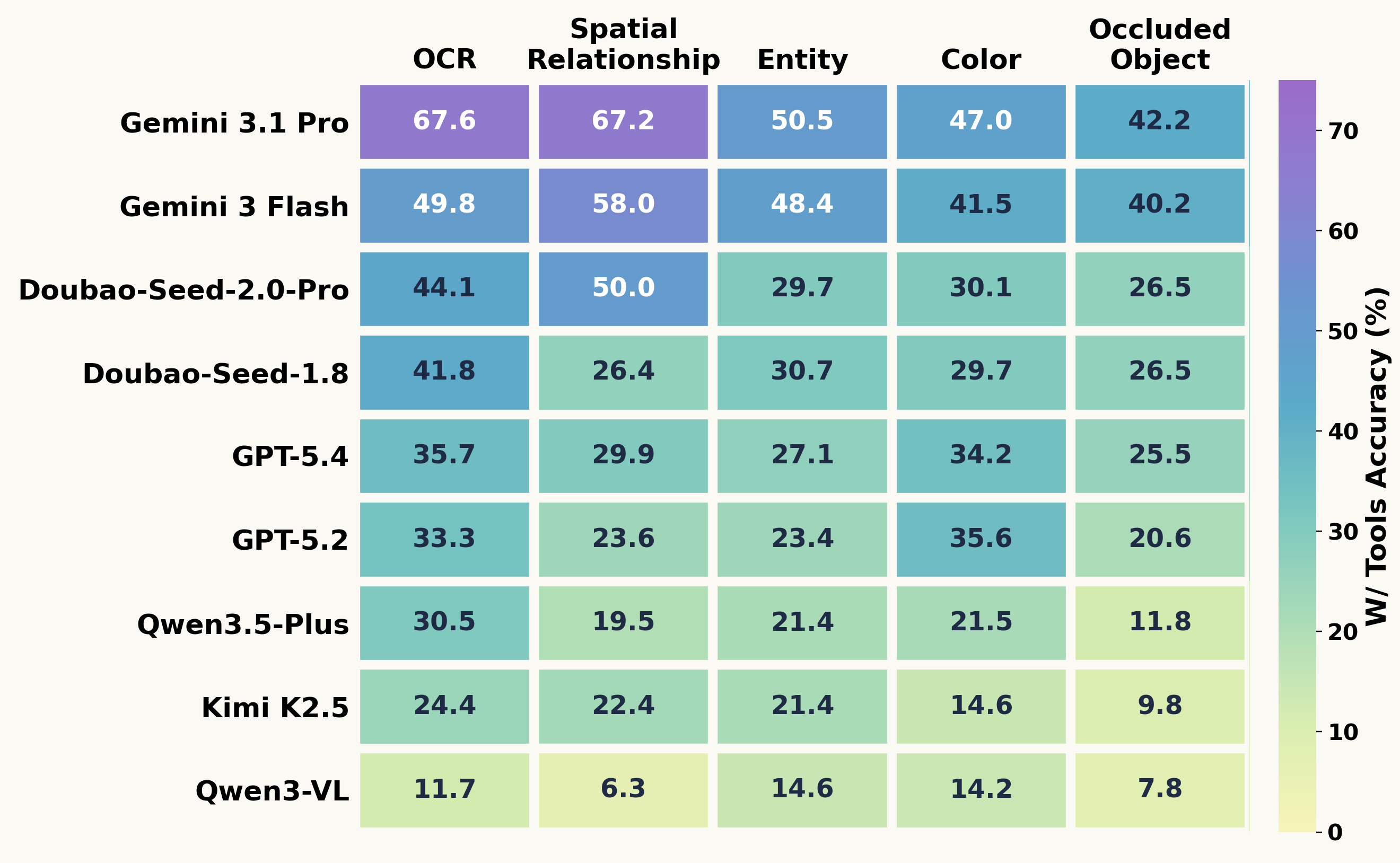}
  \end{minipage}
  \caption{Category overview of VisualNeedle. The left panel shows the data distribution across the five categories, and the right panel shows a heatmap of accuracy w/ tools for each model across the five categories. Rows correspond to models and columns correspond to categories, with colors and cell values jointly reflecting model performance across different fine-grained ability dimensions.}
  \label{fig:category_heatmap}
\end{figure}

Figure~\ref{fig:category_heatmap} further illustrates model capability distributions across the five categories. Overall, Gemini 3.1 Pro leads across all five categories, with particularly clear advantages in OCR recognition and spatial relationship. By contrast, GPT-5.4, GPT-5.2, Doubao-Seed-1.8, Qwen3.5-Plus, and other models show larger performance variation across categories: they can maintain certain levels on OCR recognition and spatial relationship, but their accuracy drops more noticeably once tasks shift toward entity recognition and occluded object recognition. Meanwhile, Kimi K2.5 and Qwen3-VL-235B-A22B-Thinking remain lower overall and show more unstable performance across categories, indicating more systematic weaknesses in fine-grained localization, local-evidence integration, and target confirmation.

\begin{table}[th]
  \centering
  \caption{Category-wise VisualNeedle results. Best Acc. denotes the highest accuracy among the 9 models in the w/ tools setting. The w/o tools and w/ tools columns report average accuracy across the 9 models in each category; Gain reports the average tool gain. All accuracies are three-run means as in Table~\ref{tab:main_results}. Because of annotation-boundary differences, one borderline sample is included in the \texttt{Entity Recognition} category.}
  \label{tab:category_results}
  \footnotesize
  \setlength{\tabcolsep}{3.5pt}
  \begin{tabular}{lcccccc}
    \toprule
    Category & \#Q & Best Model & \shortstack{Best\\Acc. (\%)} & \shortstack{w/o tools\\Avg. Acc. (\%)} & \shortstack{w/ tools\\Avg. Acc. (\%)} & \shortstack{Gain\\(pp)} \\
    \midrule
    OCR Recognition& 71 & Gemini 3.1 Pro & 67.61 & 11.11 & 37.66 & 26.55 \\
    Spatial Relationship & 58 & Gemini 3.1 Pro & 67.24 & 12.13 & 33.72 & 21.58 \\
    Entity Recognition & 64 & Gemini 3.1 Pro & 50.52 & 14.29 & 29.69 & 15.39 \\
    Color Recognition & 73 & Gemini 3.1 Pro & 47.03 & 17.86 & 29.83 & 11.97 \\
    Occluded Object Recognition & 34 & Gemini 3.1 Pro & 42.16 & 9.91 & 23.42 & 13.51 \\
    \bottomrule
  \end{tabular}
\end{table}

\section{Comparison Across Tool Configurations}

To analyze whether introducing more tools can bring additional
performance gains in VisualNeedle, we design a multi-tool comparison
experiment. Specifically, we compare the minimal Crop-only tool
configuration with a multi-tool configuration that additionally
includes enhancement and annotation operations from OpenCV. Because
the multi-tool configuration was evaluated in a single run per model,
we compare it with the first with-tools run of the corresponding
model under the Crop-only configuration in the main experiment.
Table~\ref{tab:main_results} reports the three-run means for that
main-experiment configuration. To clarify the composition of the
multi-tool setting, Table~\ref{tab:opencv_tool_taxonomy} lists the
OpenCV tools used in this experiment and their definitions.

\begin{table}[th]
  \centering
  \caption{OpenCV tools and definitions used in the multi-tool comparison experiment. We divide these tools into four groups: geometry, image enhancement, feature and structure analysis, and drawing and measurement.}
  \label{tab:opencv_tool_taxonomy}
  \footnotesize
  \setlength{\tabcolsep}{4pt}
  \begin{tabular}{p{0.27\textwidth}p{0.67\textwidth}}
    \toprule
    \textbf{Tool Name} & \textbf{Description} \\
    \midrule
    \multicolumn{2}{c}{\textbf{Geometry}} \\
    \midrule
    \textbf{Crop} & Extracts a sub-region of the image specified by a bounding box, producing a high-resolution local view for fine-grained inspection. \\
    \textbf{Resize} & Resizes an image to a target resolution. \\
    \textbf{Rotate} & Rotates the image by a specified angle. \\
    \textbf{Translate} & Shifts the image along the horizontal or vertical direction. \\
    \textbf{Flip} & Mirrors the image horizontally, vertically, or both. \\
    \textbf{Pyramid} & Performs pyramid upsampling or downsampling. \\
    \midrule
    \multicolumn{2}{c}{\textbf{Enhancement}} \\
    \midrule
    \textbf{Convert Color (Gray / HSV / LAB)} & Converts the image to another color space for downstream processing. \\
    \textbf{In-Range Color} & Filters pixels within a specified color range. \\
    \textbf{Blur} & Applies smoothing filters to suppress high-frequency noise. \\
    \textbf{Denoise} & Removes image noise while preserving salient structures. \\
    \textbf{Threshold} & Produces binary or adaptive thresholded outputs. \\
    \textbf{Morphology} & Applies erosion, dilation, opening, and closing operations. \\
    \textbf{Histogram} & Computes histogram-based image statistics and contrast enhancement. \\
    \textbf{ConvertScaleAbs} & Adjusts brightness and contrast through absolute scaling. \\
    \textbf{Inpaint} & Restores missing or corrupted regions in the image. \\
    \midrule
    \multicolumn{2}{c}{\textbf{Feature and Structure Analysis}} \\
    \midrule
    \textbf{Canny} & Detects edges using the Canny operator. \\
    \textbf{Gradients} & Computes first- or second-order image gradients. \\
    \textbf{Contours} & Extracts contour structures from binary images. \\
    \textbf{Contour Area} & Computes the area of a contour. \\
    \textbf{Arc Length} & Computes the perimeter or curve length of a contour. \\
    \textbf{Approx Poly} & Approximates a contour with a polygon of fewer vertices. \\
    \textbf{Watershed} & Segments touching regions with watershed partitioning. \\
    \textbf{GrabCut} & Extracts foreground objects via graph-cut-based segmentation. \\
    \textbf{Flood Fill} & Expands a connected region from a seed point. \\
    \textbf{Connected Components} & Computes connected components and their statistics. \\
    \textbf{Features} & Detects local keypoints and descriptors. \\
    \textbf{Hough Lines} & Detects straight lines with the Hough transform. \\
    \textbf{Hough Circles} & Detects circular patterns with the Hough transform. \\
    \textbf{Template Match} & Locates image regions matching a template. \\
    \textbf{DFT} & Converts the image to the frequency domain. \\
    \midrule
    \multicolumn{2}{c}{\textbf{Drawing and Annotation}} \\
    \midrule
    \textbf{Draw Line} & Draws a straight line on the image. \\
    \textbf{Draw Circle} & Draws a circle on the image. \\
    \bottomrule
  \end{tabular}
\end{table}

Table~\ref{tab:tool_config_compare} shows that for Gemini 3.1 Pro,
Gemini 3 Flash, and Doubao-Seed-2.0-Pro, the \texttt{crop-only}
setting reaches 55.33\%, 47.00\%, and 38.00\%, respectively, all
higher than the 53.00\%, 42.67\%, and 31.00\% achieved in the
multi-tool setting. In this single-run comparison, the results
suggest that the local high-resolution observation provided by Crop
accounts for much of the benefit of tool use, while adding more
image-processing operations provides no further gain. One possible
reason is that the core bottleneck of the benchmark lies in tiny
regions related to the answer. After localization, additional
enhancement, annotation, or transformation operations may not
provide new key information, and may instead enlarge the action
space and increase the decision burden. Thus, the minimal tool
framework centered on Crop supports effective search in this
experiment, while the more complex tool combination does not
improve accuracy for the evaluated runs.

\begin{table}[th]
  \centering
  \caption{Results of the multi-tool comparison experiment on
  VisualNeedle. Both configurations report a single run per model.
  The crop-only results use the first w/ tools run from the main
  experiment; Table~\ref{tab:main_results} reports the corresponding
  three-run means.}
  \label{tab:tool_config_compare}
  \footnotesize
  \setlength{\tabcolsep}{8pt}
  \begin{tabular}{lcc}
    \toprule
    Model
    & \shortstack{w/ tools\\(crop-only)}
    & \shortstack{w/ tools\\(multi-tool)} \\
    \midrule
    Gemini 3.1 Pro~\citep{google2026gemini31pro}
    & 55.33\% & 53.00\% \\
    Gemini 3 Flash~\citep{google2025gemini3flash}
    & 47.00\% & 42.67\% \\
    Doubao-Seed-2.0-Pro~\citep{seed2026seed2}
    & 38.00\% & 31.00\% \\
    \bottomrule
  \end{tabular}
\end{table}

\section{Additional Evaluation of Models Specialized for Visual Search}
\label{app:specialized_visual_search}

Our main experiments evaluate the visual search capabilities of
general-purpose multimodal large language models both without tools and
with access to a Crop tool. Recent work, including
DeepEyes~\citep{zheng2026deepeyes} and ACTIVE-o3~\citep{zhu2026activeo},
has further explored post-training for visual search and visual tool use
to improve models' ability to locate, crop, and use local visual evidence.
To examine whether the limitations observed in our main experiments also
arise in models optimized for these tasks, we additionally evaluate
DeepEyes and ACTIVE-o3 both without and with tools.

Both models use approximately 7B-parameter backbones. We therefore focus
on the within-model comparison between the two settings, rather than
directly comparing their scores with those of the larger general-purpose
models in the main results table. We use each model's official code and
original tool-calling pipeline and evaluate both settings under the
VisualNeedle protocol.

\begin{table}[th]
  \centering
  \caption{Performance of models specialized for visual search on VisualNeedle.}
  \label{tab:specialized_visual_search}
  \footnotesize
  \setlength{\tabcolsep}{8pt}
  \begin{tabular}{lcc}
    \toprule
    Model & w/o tools Acc. (\%) & w/ tools Acc. (\%) \\
    \midrule
    DeepEyes & 9.00 & 7.67 \\
    ACTIVE-o3 & 6.00 & 8.33 \\
    \bottomrule
  \end{tabular}
\end{table}

As shown in Table~\ref{tab:specialized_visual_search}, DeepEyes'
accuracy decreases from 9.00\% without tools to 7.67\% with tools,
while ACTIVE-o3 improves from 6.00\% to 8.33\% but remains low in
absolute terms. Under the VisualNeedle protocol, task-specific training
for visual search or crop-based tool use did not enable these two models
to reliably find and use the decisive evidence in the small regions of
information-dense scenes. These results suggest that the difficulties
observed in the main experiments cannot be attributed solely to a lack
of crop-tool-specific optimization in general-purpose models. This
conclusion is limited to the two specialized models evaluated here and
does not establish how other post-trained models would perform.

\section{Additional Controls for the Crop-Black Diagnostic}
\label{app:crop_black_controls}

We use crop-black to test whether models rely on task-relevant visual
features returned by the Crop tool during reasoning. The model still
decides when and where to request a crop, and the tool interface and
returned image dimensions remain unchanged, but each returned crop
is replaced by an all-black image. The resulting accuracy drop
indicates that the gains from standard tool use depend on the visual
features contained in the returned crops. To verify that this finding
is not specific to all-black images, we evaluate additional controls
that alter the returned visual content in different ways.

Specifically, we consider Gaussian Blur, Mosaic, Random Crop, and
Avoid-GT Random Crop. Gaussian Blur and Mosaic disrupt fine-grained
features in the returned crops while retaining coarse image content.
Random Crop substitutes randomly selected image regions for the
model-requested crops. Avoid-GT Random Crop uses the same random-crop
setting but excludes the annotated ground-truth target from the returned
crops.
Table~\ref{tab:crop_feedback_controls} reports the results.

\begin{table*}[t]
  \centering
  \caption{Accuracy (\%) under alternative visual-feedback controls.
  The w/ tools and w/o tools columns report the three-run means from
  Table~\ref{tab:main_results}. The crop-black column reproduces
  the single-run VisualNeedle results from
  Table~\ref{tab:cropblackablation}; the other diagnostic columns
  report results from their corresponding settings. Gaussian Blur
  and Mosaic use the r32 and b32 settings, respectively.}
  \label{tab:crop_feedback_controls}
  \footnotesize
  \setlength{\tabcolsep}{5pt}
  \begin{tabular}{lccccccc}
    \toprule
    Model
    & \shortstack{w/\\tools}
    & \shortstack{w/o\\tools}
    & \shortstack{Crop-\\Black}
    & \shortstack{Gaussian\\Blur}
    & Mosaic
    & \shortstack{Random\\Crop}
    & \shortstack{Avoid-GT\\Random Crop} \\
    \midrule
    GPT-5.4
    & 31.22 & 14.33 & 8.33 & 11.33 & 12.67 & 23.67 & 10.33 \\
    Gemini 3.1 Pro
    & 56.00 & 16.33 & 12.00 & 14.00 & 17.33 & 25.67 & 13.33 \\
    \bottomrule
  \end{tabular}
\end{table*}

Gaussian Blur and Mosaic substantially reduce accuracy even though
the returned images retain coarse scene content. For GPT-5.4, they
yield accuracies of 11.33\% and 12.67\%, respectively, compared with
8.33\% under crop-black and 31.22\% under standard tool use. For
Gemini 3.1 Pro, the corresponding accuracies are 14.00\% and
17.33\%, compared with 12.00\% under crop-black and 56.00\% under
standard tool use. The large drops under Gaussian Blur and Mosaic
show that the crop-black result is not specific to returning black
images: disrupting fine-grained visual features also removes much
of the benefit of tool use.

Random Crop achieves 23.67\% for GPT-5.4 and 25.67\% for Gemini
3.1 Pro, above their baselines without tools but below standard tool use.
Because a randomly sampled crop may still include the annotated
target, this residual benefit may partly come from access to
target-relevant features. Avoid-GT Random Crop tests this possibility by
excluding the target from the returned random crops. Accuracy then
falls to 10.33\% and 13.33\%, respectively, below the corresponding
baselines without tools (14.33\% and 16.33\%). Together, these controls
show that access to the tool interface or arbitrary image regions
does not reproduce the gains from model-requested real crops. They
support crop-black as an effective diagnostic ablation for testing
whether models use task-relevant intermediate visual features during
reasoning.

\section{Case Study of Search Trajectories}

\begin{figure}[th]
  \centering
  \includegraphics[width=\textwidth]{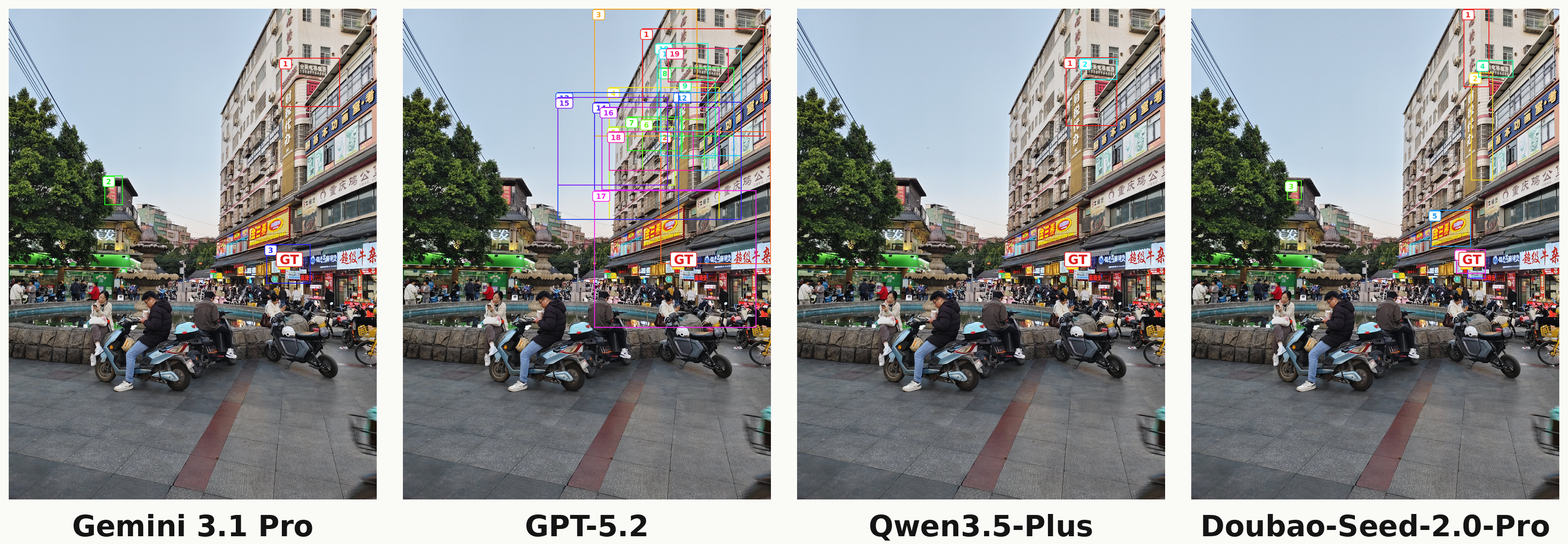}
  \caption{Comparison of four-model search trajectories on an OCR recognition example. This example requires models to locate tiny text on a red sign in a complex street scene and recognize the characters. Gold boxes denote ground-truth target regions, colored boxes denote zoom boxes at each step, and numbers denote step indices.}
  \label{fig:trajectory_case_ocr}
\end{figure}

\begin{figure}[th]
  \centering
  \includegraphics[width=\textwidth]{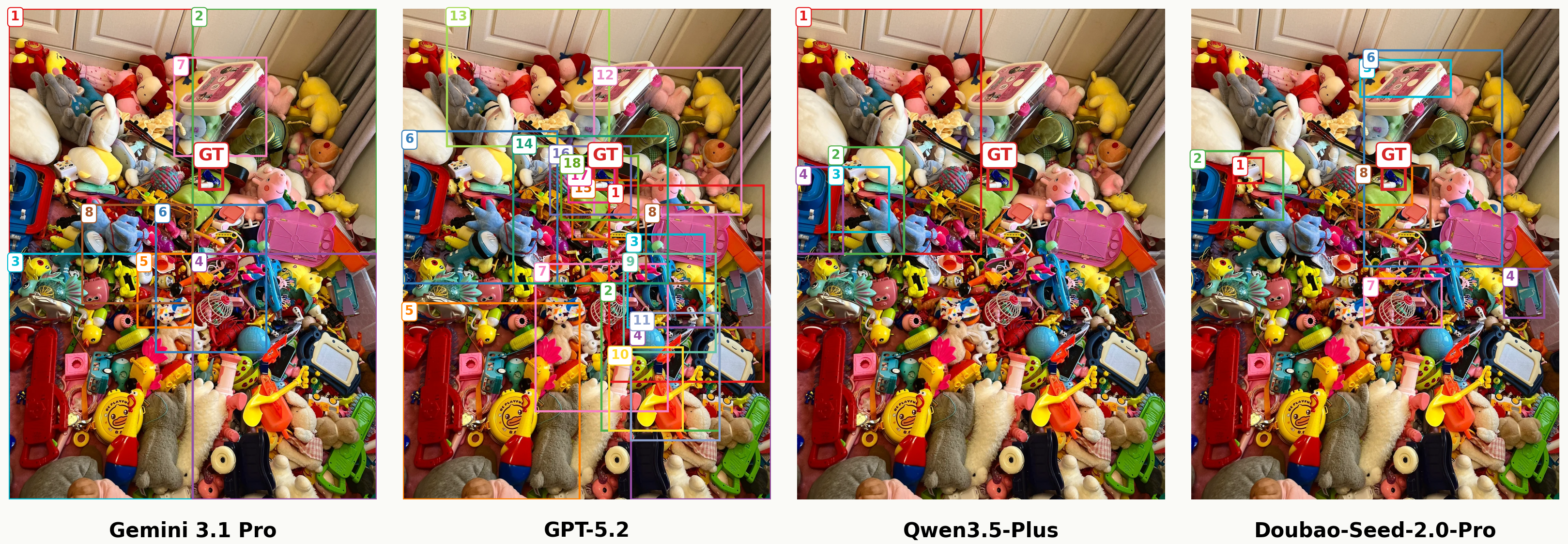}
  \caption{Comparison of four-model search trajectories on an Entity example. This example requires models to locate the blue target closest to a panda-patterned card among multiple similar small objects and complete entity confirmation. Gold boxes denote ground-truth target regions, colored boxes denote zoom boxes at each step, and numbers denote step indices.}
  \label{fig:trajectory_case_entity}
\end{figure}

\begin{figure}[th]
  \centering
  \includegraphics[width=\textwidth]{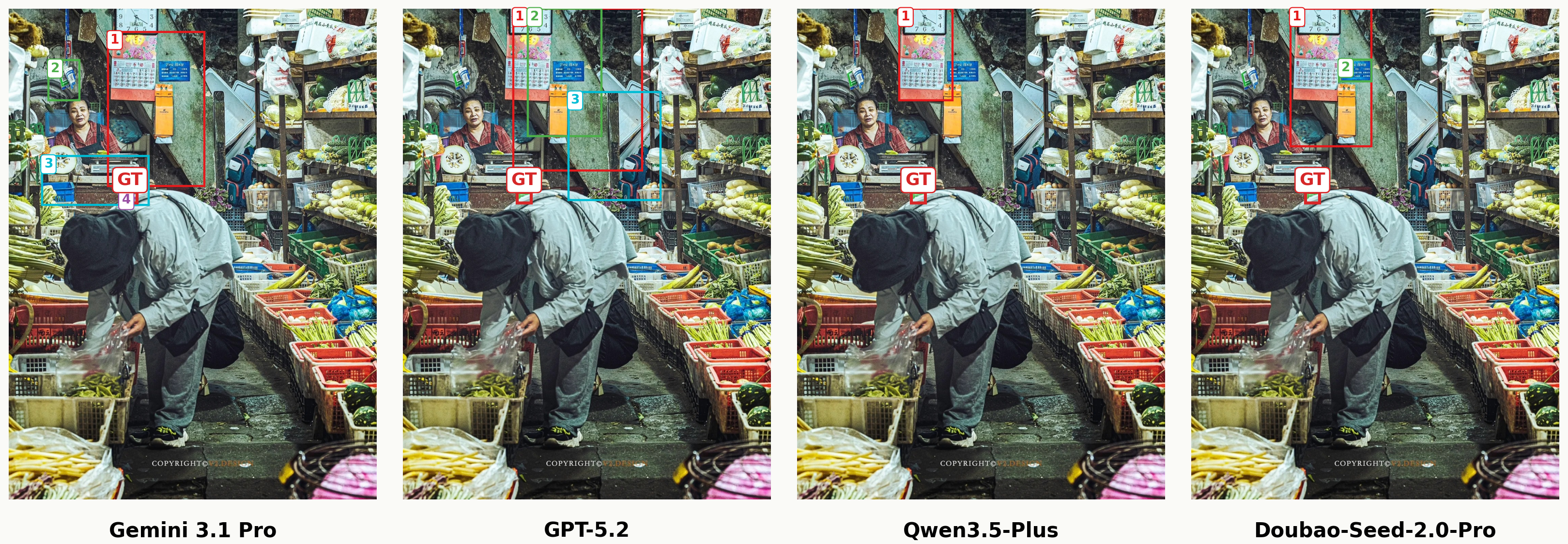}
  \caption{Comparison of four-model search trajectories on an Occluded Object example. This example requires models to locate a posted item behind a shopkeeper's wooden table under occlusion and background clutter, and then determine its function. Gold boxes denote ground-truth target regions, colored boxes denote zoom boxes at each step, and numbers denote step indices.}
  \label{fig:trajectory_case_occluded}
\end{figure}

Figures~\ref{fig:trajectory_case_ocr}, \ref{fig:trajectory_case_entity}, and \ref{fig:trajectory_case_occluded} present three representative search trajectories, corresponding to the \textbf{OCR}, \textbf{Entity}, and \textbf{Occluded Object} abilities. The OCR recognition example requires models to locate an extremely small text region in an information-dense street scene and recognize the local characters after hitting the target. The Entity example requires precise localization and confirmation among multiple visually similar candidate objects. The Occluded Object example further requires models to identify a target under partial occlusion and incomplete evidence. Together, these cases show that the difficulty of VisualNeedle lies not only in seeing the target, but also in quickly narrowing the search region in complex scenes and stably converting local visual evidence into the final answer.

In terms of model behavior, Gemini 3.1 Pro shows faster early localization and more stable convergence in all three examples. Here we say that a step \emph{hits} the target when its zoom box fully contains the ground-truth region. For instance, in the OCR recognition example, it hits the target and answers correctly at step 3, whereas GPT-5.2 does not hit the target until step 17 and still answers incorrectly. In the Entity example, Gemini 3.1 Pro first hits the target at step 2 with a coarse quadrant crop, keeps probing other regions, and answers correctly after step 8, while GPT-5.2 hits it only at step 14 and still fails to confirm it correctly. In the Occluded Object example, Gemini 3.1 Pro first hits the target at step 3, zooms in on it at step 4, and answers correctly, whereas GPT-5.2, Qwen3.5-Plus, and Doubao-Seed-2.0-Pro never truly hit the target region, indicating that only a few models can perform effective search under occlusion. Overall, GPT-5.2 often needs a longer probing process and may still fail to answer correctly even after hitting the target. Qwen3.5-Plus is more likely to stop early before entering an effective region. Doubao-Seed-2.0-Pro lies between the two, sometimes gradually approaching the target but remaining unstable in evidence integration and final confirmation.

\FloatBarrier
\section{Prompts}
\label{app:prompts}

This section summarizes the prompt templates used in our evaluation scripts.
The Crop setting does not use a separate natural-language prompt beyond the
with-tools VQA prompt. Instead, Crop is exposed through the model-facing tool
schema of \texttt{image\_zoom\_in\_tool\_reason}; the model is instructed by the
main prompt to call the available zoom/crop tool when local details are unclear.
We show the relevant templates below and omit only runtime values such as image
paths, questions, model responses, and credentials.

\begin{promptbox}{Main experiment system prompt: w/ tools VQA}
\textbf{System message.}

You are a visual question answering assistant specializing in detailed image
analysis.
You have access to the Crop tool to inspect local image regions more closely.
The tool can operate on the original input image or on a cropped image returned
by an earlier tool call.

\textbf{Structured Thinking Process.}
Follow this iterative process for each question:
\begin{enumerate}
  \item First, look closely: examine the full image and identify what
  information you need to answer the question.
  \item Identify regions of interest: if the answer depends on small or unclear
  details (text, numbers, fine print, charts, tables, or small objects),
  identify the approximate region to inspect.
  \item Use the Crop tool to inspect the relevant region. Follow the tool
  schema exactly and provide all required parameters. If cropping the original
  image or a previously returned crop, select the correct image using
  \texttt{img\_idx}. Specify the region using \texttt{bbox\_2d} in the
  coordinate format required by the tool schema, and provide a reason for the
  crop.
  \item Analyze the returned crop: carefully examine the local visual details
  and extract the information relevant to the question.
  \item Repeat if needed: you may call the Crop tool multiple times to inspect
  other regions or obtain a closer view. When cropping a previously returned
  image, select its correct \texttt{img\_idx} and specify coordinates relative
  to that image.
\end{enumerate}

\textbf{Tool-use Policy.}
Use the Crop tool whenever text, numbers, or small visual details are not clear
enough to answer confidently. Do NOT guess when a relevant region can be
inspected more closely. Avoid redundant crops, and provide all required
parameters for every tool call as specified in the tool schema.

\textbf{Answer Policy.}
After completing your analysis (including any Crop calls), output ONLY the final
answer text on the last line. Do NOT add any explanation, prefixes
(e.g., ``\zh{答案：}''), or quotes around the answer.

\end{promptbox}

\begin{promptbox}{Main experiment user prompt: w/ tools VQA}
\textbf{User message template.}

Question: \texttt{\{question\}}

Instructions:
\begin{enumerate}
  \item First, examine the full image and identify what information you need
  to answer the question.
  \item If the answer depends on text, numbers, small objects, or any detail
  that is not clearly visible, use the Crop tool to inspect the relevant
  region. You may crop the original image or an image returned by an earlier
  Crop call. Select the correct image with \texttt{img\_idx}, specify the
  region with \texttt{bbox\_2d}, and provide the required reason. Follow the
  coordinate format and other requirements in the tool schema.
  \item Examine the returned crop and repeat cropping if needed. After
  verifying the relevant details, provide your answer.
\end{enumerate}

Do NOT guess if a relevant region can be inspected more closely with Crop.
Output ONLY the final answer text on the last line.
\end{promptbox}

\begin{promptbox}{Crop tool interface}
\textbf{Tool name.} \texttt{image\_zoom\_in\_tool\_reason}

\textbf{Model-facing description.}

Zoom in on a specific region of an image. REQUIRED params: \texttt{reason}
(string), \texttt{img\_idx} (number; 0 refers to the first image visible in
this conversation, subsequent indices refer to images returned by earlier tool
calls in chronological order), \texttt{bbox\_2d} (array
\texttt{[x1,y1,x2,y2]} in 0--1000 coords). All three parameters must be
provided.

\textbf{Required fields.}\par\nopagebreak[4]

\texttt{reason}: reasoning for this zoom. Include what you learned from
previous zooms (if any), why this region needs inspection, and how this differs
from previous zooms to avoid duplication.

\texttt{img\_idx}: index of the image to reference. 0 is the first image visible
to the model in the conversation; larger indices are images returned by previous
tool calls in chronological order.

\texttt{bbox\_2d}: the bounding box of the region to zoom in, as
\texttt{[x1, y1, x2, y2]} in relative coordinates normalized to
\texttt{[0, 1000]} (not pixels), where \texttt{(x1, y1)} is the top-left corner
and \texttt{(x2, y2)} is the bottom-right corner.

\textbf{Optional field.}
\texttt{label}: a descriptive label for the target region, such as
\texttt{store\_sign} or \texttt{person\_with\_purple\_bag}.
\end{promptbox}

\begin{promptbox}{Text-only diagnostic prompt}
\textbf{System message.}

Answer the user’s question based only on the question text and your prior knowledge. You do not have access to any image.

\textbf{User message template.}

Question: \texttt{\{question\}}

You do NOT have access to any image. Based only on the question text and your
prior knowledge, give your single best guess for the answer.

Output ONLY the final answer text on the last line.

\textbf{Retry suffix if the first response is unparsable.}

IMPORTANT: Output ONLY the final answer text on the last line. Do NOT add any
explanation, prefixes (e.g., ``\zh{答案：}''), or quotes.
\end{promptbox}

\begin{promptbox}{Judge system prompt}
\textbf{System message.}

Given a question and its golden label, judge whether the model's output
correctly answers the question with a response that matches the golden label.

\textbf{Ignore these differences (format tolerant).}
Markdown formatting: \texttt{**bold**}, \texttt{*italic*}, \texttt{code}.
Quotes, brackets, and parentheses: ``\zh{答案}'' equals \zh{答案};
(\zht{工商舖}) equals \zht{工商舖}. Extra explanation or preamble
before/after the answer; whitespace and punctuation differences; case
differences such as ``hello'' = ``Hello''; simplified/traditional Chinese such
as \zh{工商铺} = \zht{工商舖}; units or measure words such as
``3\zh{个}'' = ``3'' and ``\zh{两台}'' = ``2''; digit formatting such as
``9736-8443'' = ``97368443''.

\textbf{These differences mean NO (content strict).}
Any numeric difference, such as ``3'' $\neq$ ``5'', ``\zh{五楼}''
$\neq$ ``\zh{二楼}'', or ``12'' $\neq$ ``13''. Different entities, such as
\zh{耳机} $\neq$ \zh{手机}, \zh{北京} $\neq$ \zh{上海}, or Toyota
$\neq$ Honda. Different color, such as \zh{黑色} $\neq$ \zh{白色}.
Different code/ID, such as ``E-461236'' $\neq$ ``E-461237''.

\textbf{Output.} YES or NO.

\end{promptbox}

\begin{promptbox}{Judge user prompt}
\textbf{Image-based judge user template.}

The judge receives the original image together with the following text:

Question asked about the image: \texttt{\{question\}}

Ground-truth label answer: \texttt{\{gold\_answer\}}

Model's parsed answer: \texttt{\{parsed\_answer\}} or \texttt{<NONE>}

Model's full response: \texttt{\{model\_response\}}

Based on the image, question, and the answers above, is the model's answer
correct? Output EXACTLY one word: YES or NO.
\end{promptbox}



\end{document}